\documentclass[10pt,twocolumn,letterpaper]{article}

\usepackage{cvpr}              
\definecolor{cvprblue}{rgb}{0.21,0.49,0.74}
\usepackage[pagebackref,breaklinks,colorlinks,allcolors=cvprblue]{hyperref}

\usepackage{graphicx}
\usepackage{hyperref}
\usepackage{url}
\usepackage{graphicx}
\usepackage{wrapfig}
\usepackage{booktabs}
\usepackage{tcolorbox}
\tcbuselibrary{skins, breakable}
\usepackage{multicol}

\def\paperID{7473} 
\def\confName{CVPR}
\def\confYear{2026}

\title{\emph{Kontinuous Kontext}: Continuous Strength Control \\ for Instruction-based Image Editing}

\author{
Rishubh Parihar\textsuperscript{1,3} \quad
Or Patashnik\textsuperscript{1,2} \quad
Daniil Ostashev\textsuperscript{1} \quad 
Venkatesh Babu Radhakrishnan\textsuperscript{3} \\
Daniel Cohen-Or\textsuperscript{1,2} \quad
Kuan-Chieh Jackson Wang\textsuperscript{1} \\ \\
\textsuperscript{1} Snap Research \quad
\textsuperscript{2} Tel Aviv University \quad
\textsuperscript{3} IISc Bangalore
}

\begin{document}

\twocolumn[{%
		\renewcommand\twocolumn[1][]{#1}%
		\maketitle
		\begin{center}
			\includegraphics[width=0.9\textwidth]{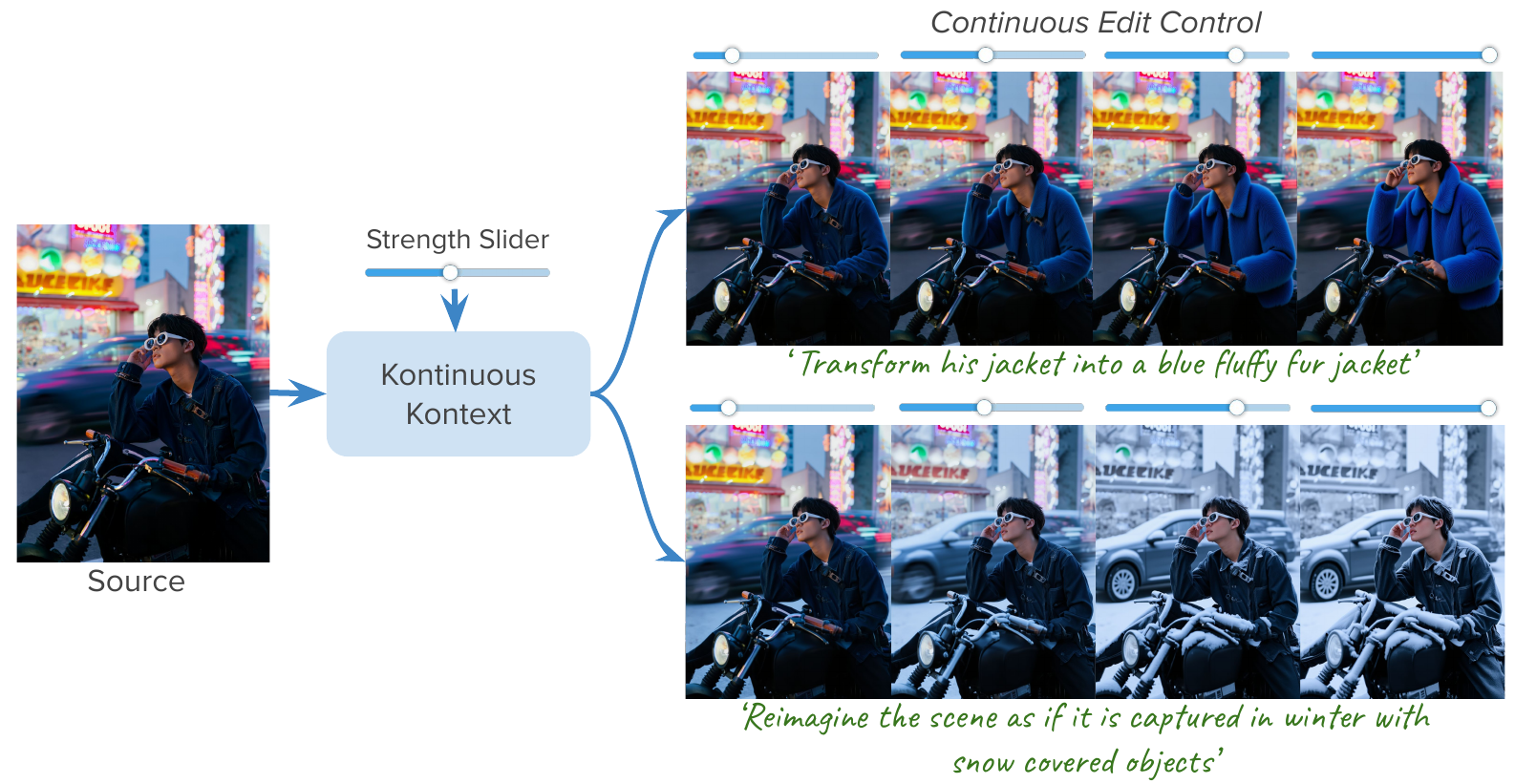}
            \vspace{-3mm} 
   \captionsetup{type=figure}
			\captionof{figure}{
               \emph{Kontinuous Kontext} produces smooth edit trajectories across diverse attributes given an image, instruction, and an edit scalar strength. Unlike prior methods that require attribute-specific training, ours is a unified approach to enable fine-grained control.}
			\label{fig:teaser}
		\end{center}
	}]

\maketitle
\begin{abstract}
Instruction-based image editing offers a powerful and intuitive way to manipulate images through natural language. Yet, relying solely on text instructions limits fine-grained control over the extent of edits. We introduce Kontinuous Kontext, an instruction-driven editing model that provides a new dimension of control over edit strength, enabling users to adjust edits gradually from no change to a fully realized result in a smooth and continuous manner. Kontinuous Kontext extends a state-of-the-art image editing model to accept an additional input, a scalar edit strength, which is then paired with the edit instruction, enabling explicit control over the extent of the edit. To inject this scalar information, we train a lightweight projector network that maps the input scalar and the edit instruction to coefficients in the model's modulation space. For training our model, we synthesize a diverse dataset of image-edit-instruction-strength quadruplets using existing generative models, followed by a filtering stage to ensure quality and consistency. Kontinuous Kontext provides a unified approach for fine-grained control over edit strength for instruction driven editing from subtle to strong across diverse operations such as stylization, attribute, material, background, and shape changes, without requiring attribute-specific training.

\end{abstract}

\vspace{-2mm}
\section{Introduction}

\begin{figure*}
    \centering
    \includegraphics[width=0.90\linewidth] {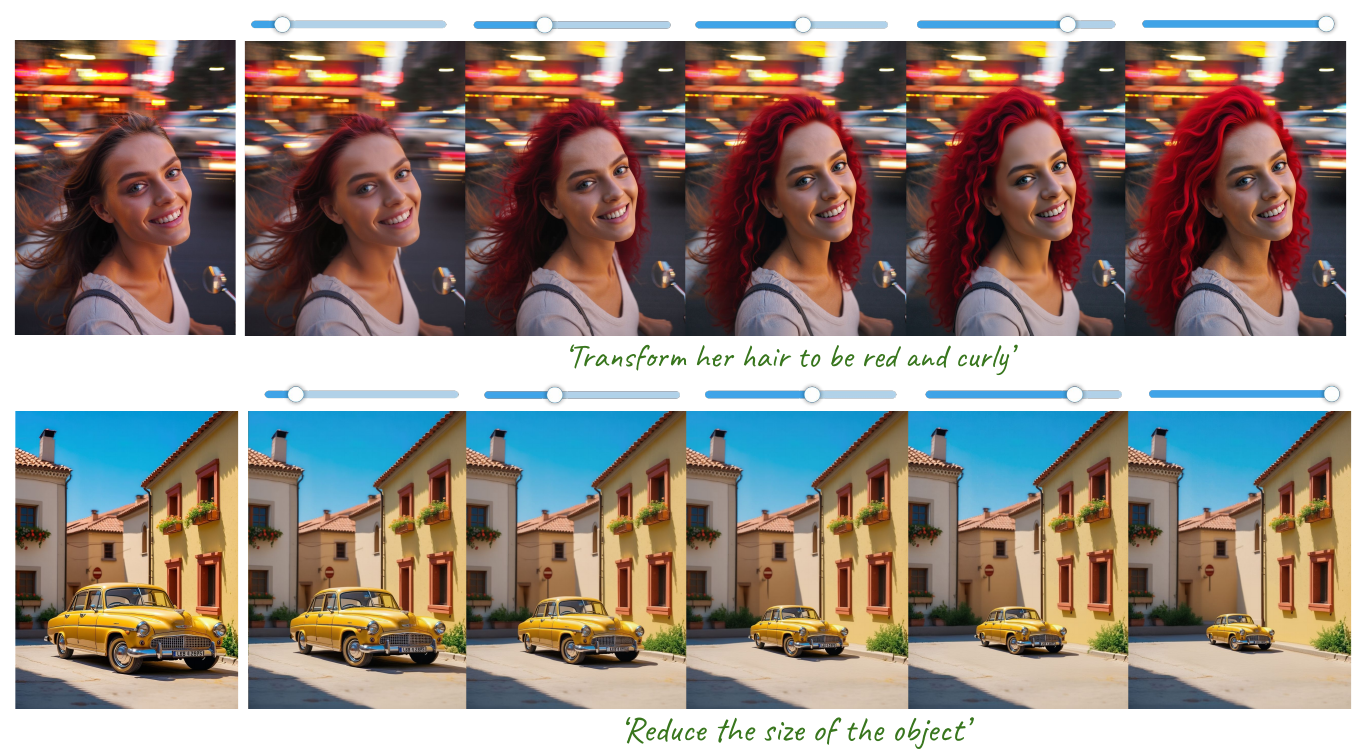}
    \vspace{-3mm}
    \caption{\emph{Kontinuous Kontext} enables finer control across diverse instruction-driven image edits. It can do simultaneous changes in attributes, hair color, and structure, and geometric edits such as changing the size of the car.}
    \vspace{-5mm}
    \label{fig:results2} 
\end{figure*}

The advent of large-scale text-to-image generative models~\citep{ho2020denoisingdiffusionprobabilisticmodels, song2022denoisingdiffusionimplicitmodels, rombach2022high} has enabled phenomenal progress in instruction-driven image editing, allowing users to perform a broad range of edits through natural language instructions~\citep{hertz2022prompt, brooks2023instructpix2pix, batifol2025flux}.
With a single prompt (e.g., \textit{`make the person old'}), these models can change style, modify object appearance or shape, and add or remove objects. 
While text is an intuitive interface for specifying editing goals, it is also a coarse modality: it conveys what change to make but not to what extent. As a result, users lack fine-grained control over the edit strength (e.g., adjusting the degree of \textit{`oldness'} in a portrait). This limitation poses a central challenge for achieving precise and controllable image manipulation.

To address this challenge, prior work has explored continuous control for image manipulation, ranging from GAN-based latent space editing~\citep{shen2020interpreting, härkönen2020ganspace, abdal2021styleflow, Patashnik_2021_ICCV} to diffusion-based methods 
that rely on specialized per-attribute modules~\citep{cheng2025marble, gandikota2024concept, sharma2024alchemist}. While these approaches demonstrate the appeal of continuous editing, they are often restricted to narrow domains or require dedicated training for each attribute. This leaves open the need for a unified method that enables continuous control across diverse types of edits without the burden of training per-attribute models.

In this work, we introduce \emph{Kontinuous Kontext}, an instruction-driven image editing model that introduces a new dimension of control, enabling continuous adjustment of edit strength across diverse edit categories. Rather than being limited to a binary “before/after” operation, our approach enables smooth traversal between no edit and a fully realized edit, turning coarse instructions into rich, tunable controls. For example, users can gradually change the extent of stylization or intensity of snowfall (Fig.~\ref{fig:teaser}), as well as perform local edits with finer control including attribute edits such as hair shape and color, or object size (Fig.~\ref{fig:results2}). By transforming discrete instructions into continuous editing trajectories, our method bridges the gap between intuitive text prompts and fine-grained user control, offering a level of precision unattainable with text alone.

We realize this new dimension of control by augmenting an existing instruction-based image editing model with an additional input scalar that specifies edit strength.
Specifically, we build on Flux Kontext~\citep{batifol2025flux}, a state-of-the-art instruction-driven image editing model and condition it with the strength scalar via a lightweight projector network. The projector takes as input the scalar value together with the edit instruction embeddings and outputs coefficients calibrated to the specific edit instruction.
These coefficients operate in the model's modulation space~\citep{garibi2025tokenverse, dalva2024fluxspace}, where they modulate the text tokens, effectively refining the edit instruction to reflect the desired strength. 

Training the projector requires data consisting of source image, edit instruction, edit image, and annotations of edit strengths, which is not readily available for real images. To overcome this limitation, we synthesize such tuples using existing generative techniques. Specifically, we first use an LVLM~\citep{Qwen2.5-VL} to generate diverse, image-specific edit instructions. Next, we apply instruction-driven image editing model~\cite{batifol2025flux} to produce edited images from the source images and the synthesized instructions. Finally, we use a diffusion based image morphing model~\citep{cao2025freemorph} to generate intermediate edits at varying strengths. 
The synthesized data, however, often provides noisy supervision, where the sequences are not smooth or the intermediate images have artifacts or deviate too far from the endpoints. To address this, we apply filtering based on identity preservation of input images and smoothness of the edit transitions to obtain clean, reliable training data. In addition, the scale and diversity of the dataset helps mitigate remaining inaccuracies and outliers. Notably, we find that even when trained on this high quality filtered but moderately sized dataset, our method generalizes strongly across diverse editing categories.





\vspace{-0.5mm}
Extensive experiments across a broad spectrum of instruction-driven editing tasks show that \emph{Kontinuous Kontext} provides rich, diverse, and finely controlled edits. It enables precise strength control for local edits such as attribute, material or appearance changes, global transformations such as style, environment or lighting changes, and even challenging geometric edits like shape morphing. Notably, it generalizes beyond its training categories to unseen cases such as facial attributes and body shape changes. These findings establish our approach as a powerful, general framework for continuous instruction-driven image editing, opening new directions for fine-grained and controllable visual editing.

\section{Related Works}
\vspace{-1mm}

\paragraph{Instruction-driven Image Editing.} The advancements of scalable visual generative models ~\citep{esser2024scaling,podell2023sdxl,ramesh2022hierarchical,wu2025qwen,rombach2022high} trained on internet-scale image-text pairs have fueled a wide range of image editing applications. Instruction-based image editing, introduced by Instruct-Pix2Pix~\citep{brooks2023instructpix2pix} enables editing images with text instructions. To this end, they curated a synthetic dataset of image-edit pairs generated using Prompt2Prompt~\citep{hertz2022prompt}, with corresponding editing instructions generated by an LLM, and fine-tuned the Stable Diffusion model~\citep{rombach2022high} for instruction-driven editing. Subsequently, many works ~\citep{sheynin2024emu, zhang2025context,zhang2024hive} have improved the dataset curation pipeline and model architecture, leading to stronger instruction-following ability. More recent approaches train large unified models for both generation and editing ~\citep{batifol2025flux,wu2025qwen,wu2025omnigen2,xiao2025omnigen}. These models are capable of performing diverse editing tasks such as personalization, scene composition, and instruction-based editing. Despite their remarkable general-purpose editing capabilities, these models lack control over the extent of the edit, limiting their applicability for users who require fine-grained adjustments.

\vspace{-2mm}
\paragraph{Discovering Continuous Control in Generative Models.}
A common approach to achieve control over edit strength is through traversals in latent spaces. 
In GANs and VAEs, compressed latent representations capture rich semantics, enabling the discovery of directions that correspond to semantic attributes~\citep{karras2019style, härkönen2020ganspace,hou2017deep,higgins2017beta}. Numerous traversal methods have been developed to leverage these directions for fine-grained attribute manipulation~\citep{shen2020interpreting,abdal2021styleflow,Patashnik_2021_ICCV}. However, such methods remain restricted to narrow domains.
Extending the idea of latent space traversal to diffusion models is challenging, as the denoising network does not naturally provide a compact latent space~\citep{kwon2022diffusion}, text embeddings are not smooth~\citep{hertz2022prompt}, and LoRA-based weight interpolations~\citep{gandikota2024concept,gandikota2025sliderspace, dravid2024interpreting} remain computationally expensive and concept-specific.
These approaches all rely on discovering latent or weight-space directions with continuous variation. In contrast, we augment the instruction mechanism with a new control dimension, enabling smooth adjustment of any attribute the model can already edit. Hence, our model does not require any additional training for specific attributes.

\vspace{-4mm}
\paragraph{Adding Continuous Control for Image Editing.}
Another set of works introduces continuous control in diffusion models by either fine-tuning the model itself or training auxiliary encoders that modify its inputs. Some works~\citep{sharma2024alchemist,cheng2025marble,magar2025lightlab} generate synthetic data with varying material or illumination properties using rendering engines and fine-tune diffusion models for continuous control over these attributes. Others train encoders to predict new token embeddings injected into the text embedding space, enabling control over 3D properties such as orientation, illumination, and shadows~\citep{cheng2024learning,parihar2025compass,burgess2024viewpoint}. A further line of work trains adapters that connect the continuous latent spaces of GANs with the stronger generative capabilities of diffusion models, specifically for face attribute editing~\citep{parihar2024precisecontrol,li2024stylegan}. Despite their effectiveness, methods across these directions remain limited to a single attribute or object category.

\begin{figure*}
    \centering
    \includegraphics[width=\linewidth]{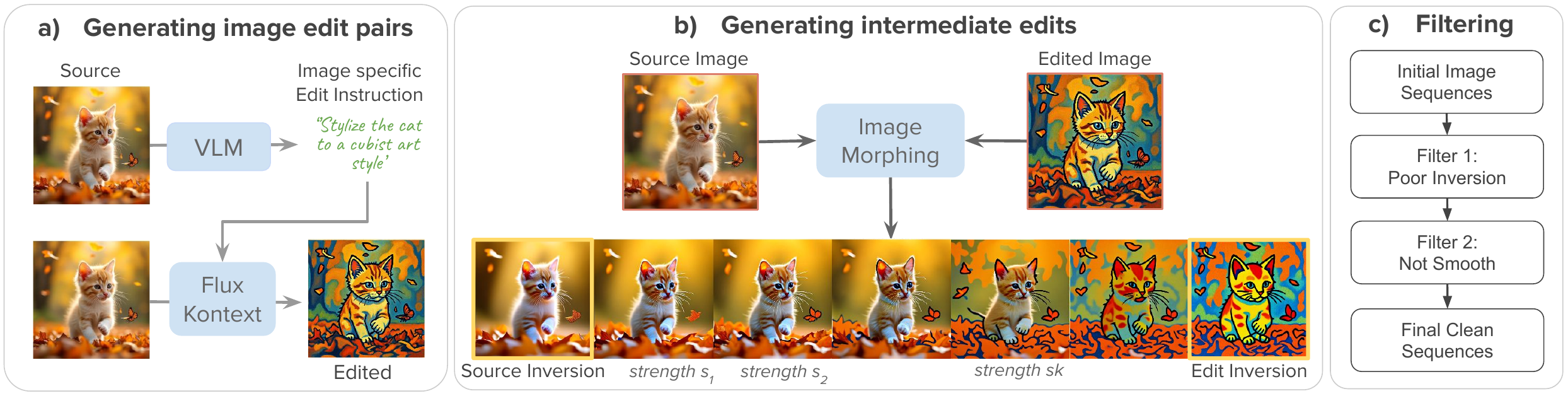} 
    \vspace{-5mm}
    \caption{\textbf{Data generation.} Our pipeline consists of three steps: (a) We generate an edit instruction for each source image using a pretrained VLM, then apply Flux Kontext, an instruction-driven editing model, to produce a full-strength edit. (b) We synthesize intermediate-strength edits using a diffusion-based morphing method~\citep{cao2025freemorph}, which inverts both the source and edited images into the diffusion latent space and interpolates their features. (c) To compensate for inconsistencies in the morphing sequence (Fig.~\ref{fig:data-filtering}), we filter the samples based on the inversion quality and uniformity of the sequence.
    }
    \vspace{-6mm}
    \label{fig:dataset}
\end{figure*} 


\vspace{-4mm}
\paragraph{Image Interpolation.}
A promising baseline strategy for continuous control in image editing could be to generate the edited image with instruction and then generate intermediate images between the source and the edit. Diffusion-based morphing methods~\citep{cao2025freemorph,zhang2024diffmorpher} aim to generate intermediate transitions by interpolating in the diffusion feature space, under the assumption that this space is semantically smooth. While this assumption holds in some cases, the space is not robust to outliers and often produces artifacts in intermediate morphs, such as missing objects or blurred scene content (Fig.~\ref{fig:data-filtering}). Another option is to adapt video inbetweening models~\citep{wan2025,zhu2025generative,wanggenerative} to synthesize intermediate frames as continuous edits. However, as these models are trained on natural videos, they produce abrupt transitions for imaginative edits such as stylization or attribute changes, and their outputs frequently exhibit motion blur, making them unsuitable for high-quality image editing.

   

\vspace{-1mm}
\section{Method}
\vspace{-2mm}
We extend instruction-driven image editing by introducing a new dimension of control: continuous adjustment of edit strength. Our approach has two key stages. First, we generate a diverse synthetic dataset of paired examples consisting of source images, edited images, edit instructions, and continuous strength values (Sec.~\ref{subsec:dataset}). Second, we propose a simple yet effective approach: fine-tuning a modified instruction-driven editing model that accepts a scalar strength input alongside the edit instruction, enabling smooth and continuous control over the edit (Sec.~\ref{subsec:kontmodel}).

\vspace{1mm}
\subsection{Dataset}
\label{subsec:dataset}
\vspace{-1mm}
Our method utilizes a dataset of tuples $(x, e, s, y_s)$, where $x$ is a source image, $e$ is an edit instruction, $s$ is an edit strength, and $y_s$ is the corresponding target edit. Since collecting real data with multiple strength levels is challenging, we curate a synthetic dataset using pretrained generative models. An ideal dataset should ensure that each edited image $y_s$ faithfully follows the instruction $e$, and exhibit smooth transformations as the strength $s$ increases. To this end, 
we design a three step data generation process: (i) generate a full-strength edit using an existing instruction-driven editing model, (ii) interpolate between the source and the full-strength edit to produce intermediate-strength variations, and (iii) filtering poor-quality data samples. 

\vspace{-3mm}
\paragraph{Generating Image Edit Pairs.} 
We begin by sampling $110K$ images of diverse objects and scenes across different background and environment conditions from the Subject200K dataset~\citep{tan2024ominicontrol}. For each image, we generate an edit instruction using Qwen LVLM~\citep{Qwen2.5-VL}, covering a diverse category of continuous editing operations (Fig.~\ref{fig:dataset}a). We categorize edits into global scene edits (\textit{stylization}, \textit{scene reimagination}, and \textit{environment change}) and local object-specific edits (\textit{material} and \textit{appearance editing}, \textit{attribute modification}, and \textit{shape morphing}), also shown in appendix Fig.\textcolor{cvprblue}{2}. We define a fixed template system prompt for each subcategory. Additionally, we generate $n$ in-context examples using GPT4 for each of the subcategories. During instruction generation, we randomly sample from these category-specific in-context examples to guide the VLM in generating diverse instructions. The source image and its corresponding instruction are then used to produce a full-strength edit ($y^*$) with Flux Kontext~\citep{batifol2025flux}. Generating the edit from Flux Kontext ensures consistency with the base model’s output distribution. Further details of the prompts and additional samples are provided in the appendix Sec.\textcolor{cvprblue}{A.2}.

\vspace{-4mm}
\paragraph{Generating Edits with Intermediate Strengths.} 
We generate intermediate edits by synthesizing smooth transitions between the source image $x$ and the full-strength edit $y^*$ generated by Flux Kontext. We define a discrete set of $N{+}1$ edit strengths $\{s_i=i/N \mid i=0,\ldots,N\}$ uniformly sampled within the normalized range $[0,1]$. Here, $s_0=0$ corresponds to the unedited source, $s_N=1$ corresponds to the full edit $y^*$, and the intermediate values $s_i$ for $1 \leq i \leq N{-}1$ represent proportionally graded changes. Given the source and edited images, we use off-the-shelf diffusion based image morphing, Freemorph~\citep{cao2025freemorph} to generate the intermediate images $y_{s_i}$, which we treat as edits at the corresponding strengths $s_i$. Freemorph first inverts the two end point images into the latent space of pretrained diffusion model. Next, it performs guided spherical interpolation between their self-attention maps during denoising to produce intermediate morphs. This yields perceptually monotone transitions that interpolate between the two images (Fig.\ref{fig:dataset}b). We use $N=6$ as  prescribed in Freemorph~\citep{cao2025freemorph}.

\begin{figure}[h]
    \centering
    \vspace{-2mm}
    \includegraphics[width=\linewidth]{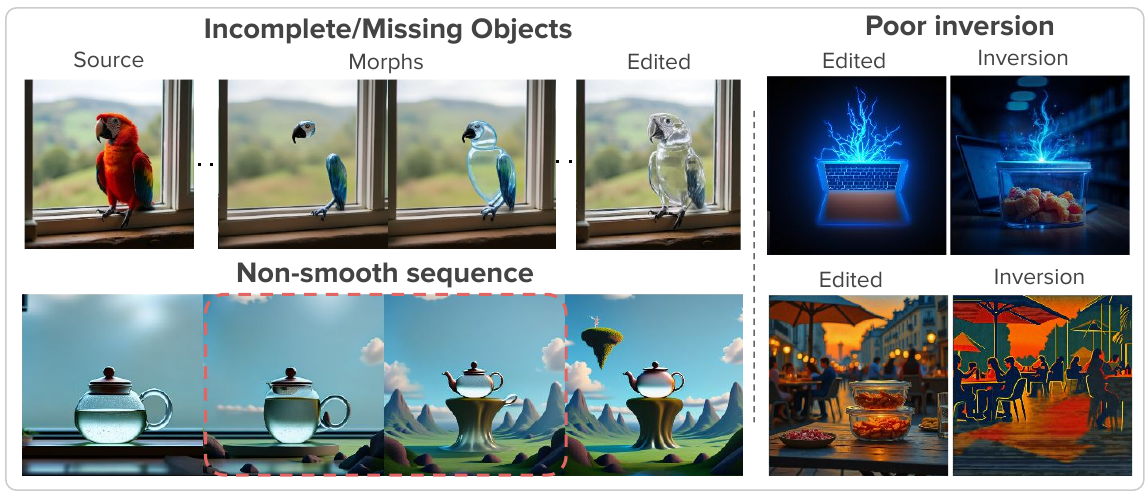}
    \vspace{-7mm}
    \caption{\small Generating intermediate images with Freemorph can introduce inconsistencies such as incomplete objects, abrupt jumps, or errors from diffusion inversion. We filter such cases to obtain a clean dataset with smooth trajectories.} 
    \vspace{-3mm}
    \label{fig:data-filtering}
\end{figure}

We observe that Freemorph has two key limitations. First, its latent space is not semantically smooth, often producing unnatural intermediate images, artifacts with incomplete objects (Fig.~\ref{fig:data-filtering}) and abrupt transitions for large edit transformations. More broadly, as an inference-time heuristic, Freemorph lacks robustness, which further contributes to the errors. To address these issues, we employ an extensive data filtering pipeline. Second, since Freemorph relies on diffusion inversion, it introduces reconstruction errors in the source and edited images during inversion, which makes the intermediate images inconsistent (Fig.~\ref{fig:dataset}b). We fix this by replacing the original endpoints with their reconstructions, ensuring consistency with the intermediate morphs. 

\vspace{-1mm}
\paragraph{Data Filtering.} 
While effective, the above data generation pipeline is prone to errors from the underlying generative models (Fig.~\ref{fig:data-filtering}), making filtering essential to eliminate inconsistent samples. To filter out samples with non-smooth edit trajectories, we quantify the uniformity of the edit trajectory and threshold on this score. 
For a training sample $(x,e,s,y_s)$, the extent of change between the source $x$ and edit $y_s$ should scale with the edit strength $s$. Equivalently, the distance between adjacent images in the sequence should remain consistent. 
We define the sequence of deltas as
$D = \{d_{0,1}, d_{1,2}, \ldots, d_{N-1,N}\}$, where $d_{i,i+1}$ is the distance between image $y_i$ and $y_{i+1}$ and measure its uniformity via the KL-divergence from a discrete uniform distribution. 

In addition to non-uniform trajectories we observe for stronger edits, the diffusion inversion step in Freemorph can drastically alter the edited image (Fig.~\ref{fig:data-filtering}). We discard such cases by thresholding the image distance between the edit and its inversion. Similarly, in some cases Flux Kontext fails to perform the edit and instead reproduces the input with minimal changes; we filter out such examples by computing image distance between the source and edited images. We used LPIPS~\citep{zhang2018unreasonable} to compute the image distance in all the filtering criteria. After filtering, our dataset is reduced from $110K$ to $64K$ high-quality, smooth and, accurate edit trajectories. We also generate a synthetic dataset of $10K$ samples by varying object size through zooming.

\subsection{Kontinuous Kontext}  
\label{subsec:kontmodel}

\begin{figure}[h]
    \centering
    \vspace{-2mm}
    \includegraphics[width=\linewidth]{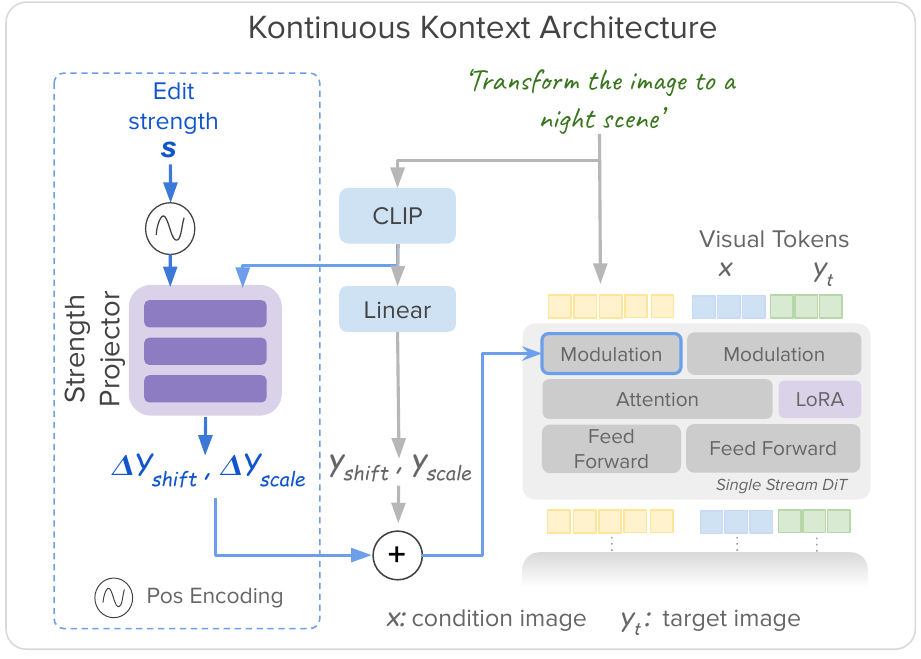}
    \vspace{-8mm}
    \caption{We design a lightweight projector network that maps a scalar edit strength $s$ to offsets of the text modulation parameters, enabling precise control over edit strength.}
    \vspace{-2mm}
    \label{fig:model-architecture}
\end{figure}

\noindent 
\textbf{Preliminaries.} Our model builds on Flux Kontext~\citep{batifol2025flux}, a DiT-based image editing model. It takes a source image and instructions and outputs the edited image. The design follows Flux~\citep{flux}, where image and text are encoded as tokens and processed through visual and textual attention streams. Flux Kontext extends this by encoding the source (context) image with the Flux encoder, then concatenating the source tokens ($x$) with the noised target tokens ($y_t$), which are jointly processed in the visual stream (Fig.~\ref{fig:model-architecture}). As in Flux, a pooled embedding of the edit instruction is fused with the timestep embedding to predict separate modulation parameters for both textual and visual tokens.

\vspace{-5mm}
\paragraph{Conditioning on edit strength.}  
Our goal is to inject the scalar edit strength into the instruction-driven Flux Kontext model~\citep{batifol2025flux}. Intuitively, edit strength can be viewed as an attribute of the instruction itself, which suggests representing it as an additional token in the text token sequence. However, our early experiments revealed that the text embedding space is not a smooth latent space for strength control, often producing abrupt transitions between adjacent edit strengths (appendix Fig.\textcolor{cvprblue}{8}). Recent works~\citep{garibi2025tokenverse, dalva2024fluxspace} have shown that the modulation space of DiT models is highly disentangled and enables fine-grained control of attributes in text-to-image generation. In particular, object-specific attributes can be modified by adjusting the modulation parameters of the corresponding word in the text prompt~\citep{garibi2025tokenverse}. Conditioning in the modulation space provides global conditioning that is structurally equivalent to widely used AdaLN~\cite{adaln}. 

\begin{figure}[h]
    \centering
    \vspace{-4mm}
    \includegraphics[width=0.95\linewidth]{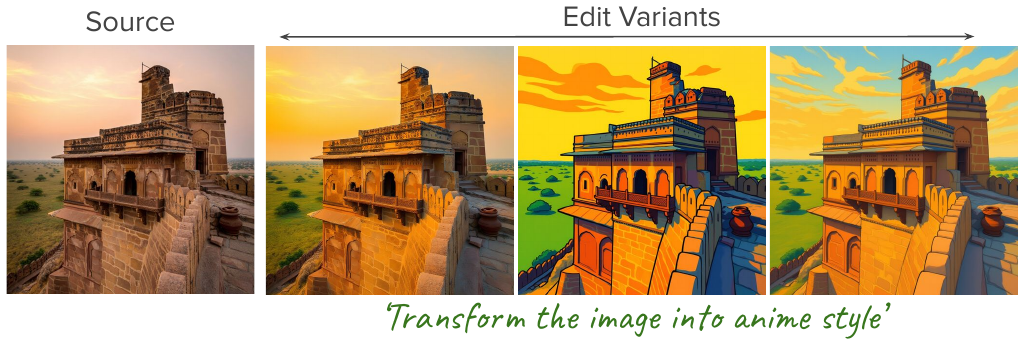}
    \vspace{-4mm}
    \caption{In a simple experiment, we scale the text-token modulation parameters in Flux Kontext with a scalar to generate edit variations. This perturbation produces edits of varying strengths, revealing that modulation parameters can govern edit strength.}
    \vspace{-4mm}
    \label{fig:motivation}
\end{figure}

We find that the modulation space of instruction-driven image editing models allows control over edit strength. In a simple experiment, we scaled the modulation parameters of the text tokens with a scalar $v \in (0.5, 2.0)$ and generated multiple edits of the same image and instruction. As shown in Fig.~\ref{fig:motivation} \& appendix Sec.\textcolor{cvprblue}{A.4.1}, perturbing the modulation parameters produce edits of varying strength, while preserving models prior of preserving image identity. Building on this insight, we inject edit-strength information into the network through the modulation parameters of the text tokens. Specifically, we design a strength projector network that maps the input scalar strength value to offsets of the original text modulation parameters, enabling appropriate adjustments for continuous control of edit strength. 

\begin{figure}[t]
    \centering
    \vspace{-2mm}
    \includegraphics[width=\linewidth]{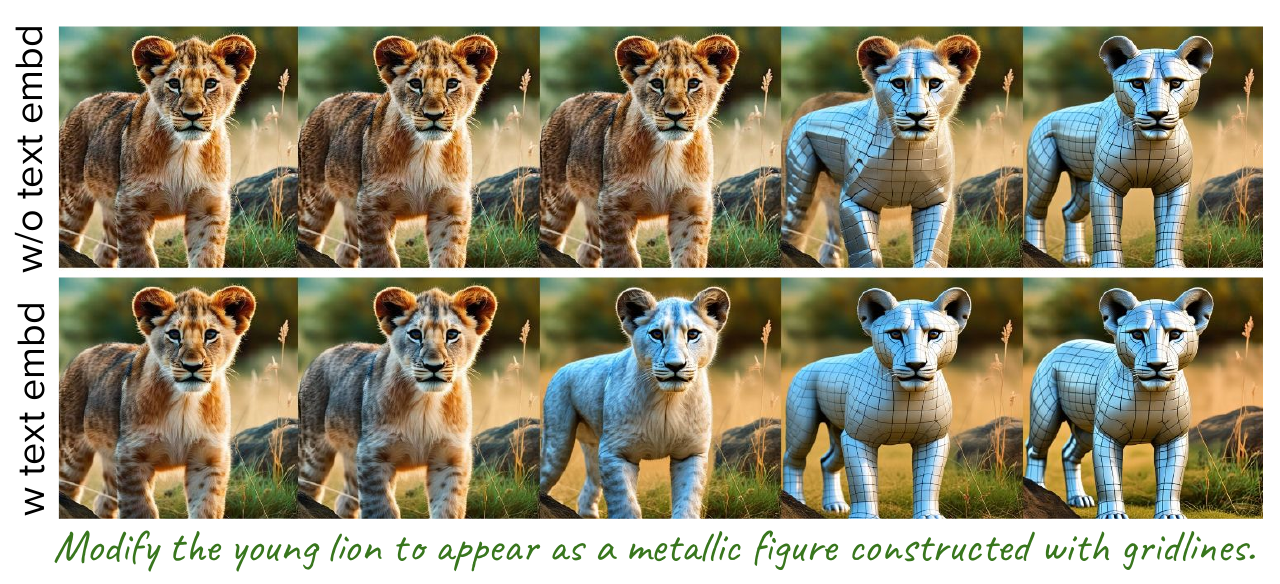}
    \vspace{-8mm}
    \caption{\small Adding text embeddings into the slider projector improves smoothness of edit transitions.}
    \vspace{-6mm}
\label{fig:clip-ablate}
\end{figure}

\vspace{-4mm}
\noindent 
\paragraph{Strength Projector} is a small MLP that maps the scalar edit strength $s \in (0,1)$ into the offsets $[\Delta y_{shift},\Delta y_{scale}]$ to the modulation parameters of the text tokens $[y_{shift}, y_{scale}]$ as shown in Fig.~\ref{fig:model-architecture}. A direct implementation of this projector would predict identical offsets for all edits at a given strength, ignoring the type of edit. This leads to uncaliberated edits resulting in sudden jumps in edits. For example,

\noindent 

\noindent 
as shown in Fig.\ref{fig:clip-ablate}, for material editing, the model generates sudden transitions. To overcome this limitation, we provide the pooled CLIP text embedding as an additional input, allowing the predicted modulation parameters to depend on the instruction. This results in calibrated modulations that enable smooth, continuous control across diverse edit categories. More details are in appendix Sec.\textcolor{cvprblue}{A.3}.

\begin{figure*}[t]
    \centering
    \vspace{-4mm}
    \includegraphics[width=0.85\linewidth]{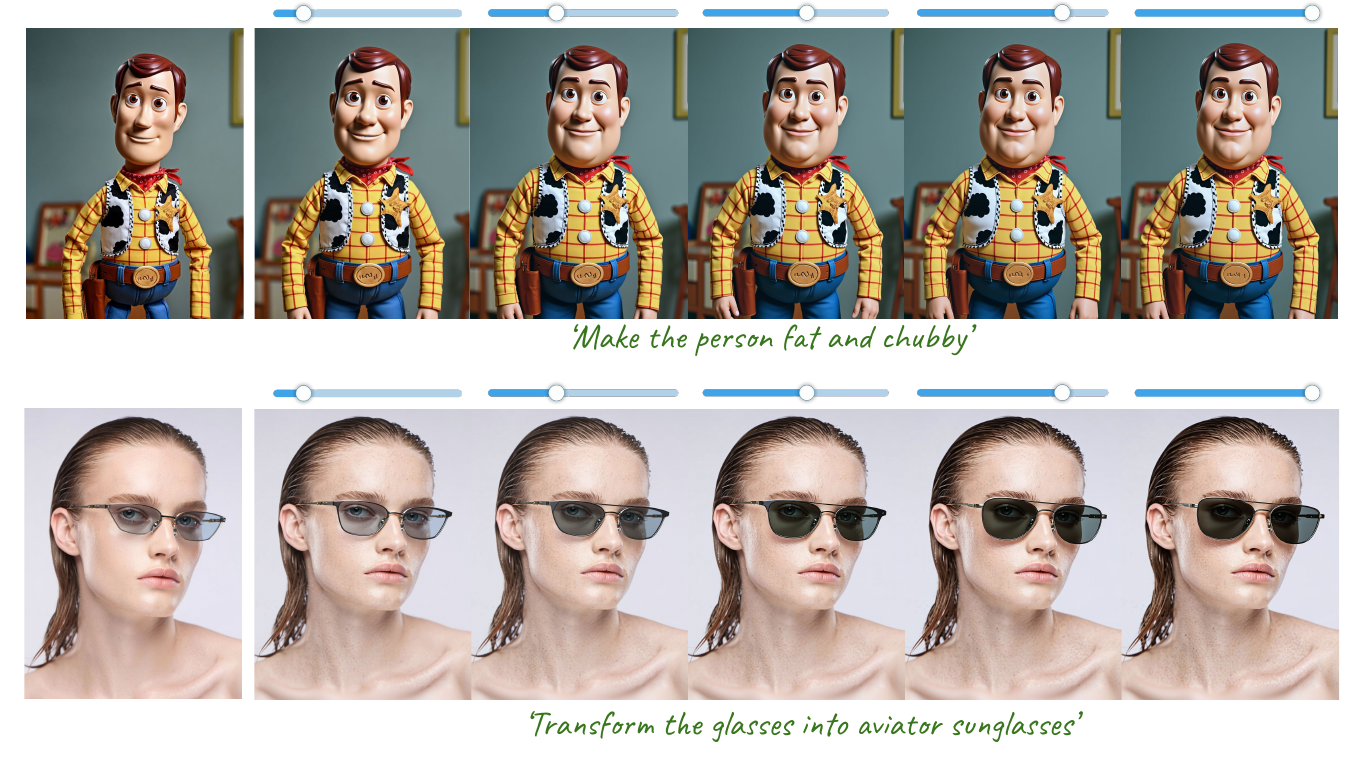}
    \vspace{-5mm}
    \caption{
    Our model achieves smooth geometric transformations, from body-shape edits to seamless eyeglass shape-color blending.
    }
    \vspace{-5mm}
    \label{fig:results-ours}
\end{figure*} 

\vspace{-2mm}
\paragraph{Training.} We train our model on the curated dataset (Sec.~\ref{subsec:dataset}) by sampling paired data consisting of source image $x$, edit instruction $e$, edit strength $s$, and target edit $y_s$. Trainable parameters include LoRA for the attention projection matrices of the Flux Kontext model, along with the projector network. Concretely, a data sample ($x, e, s, y_s$) and a diffusion timestep $t$, we optimize the model using the standard flow matching loss:

\vspace{-5mm}
\begin{equation*}
\mathcal{L}_{\theta} = \mathbb{E}_{t \sim p(t),x,e, s, y_s} 
\left[ \left\| v_{\theta}(y_s^t, t, e, x, s) - (\epsilon - x) \right\|_2^2 \right],
\end{equation*}
\vspace{-5mm}

\noindent 
where $y_s^t$ is the interpolated latent between $y_s$ and Gaussian noise $\epsilon \sim \mathcal{N}(0,1)$, defined as $y_s^t$ = $(1 - t)y_s + t\epsilon$. 
$v_\theta$ is the Kontinuous Kontext model. As a regularization, we randomly drop the slider conditioning with probability $0.1$. More details about training are in appendix Sec.\textcolor{cvprblue}{A.1}.  



\vspace{-1mm}
\section{Experiments} 
\label{sec:experiments}
\vspace{1mm}


\vspace{-2mm}
\paragraph{Evaluation Benchmark.}
We use a standard image editing benchmark, PIEbench~\citep{ju2023direct}, that consists of diverse and challenging instruction-driven image editing test examples. The benchmark consists of edits from the following editing categories: change object, add/remove object, change pose, change color, change material, change background and change style. We remove the add/remove category as it is not a continuous edit. The instructions involved challenging edits that often have two-three edits in one prompt (e.g., \textit{`transform the dog into a brown german shepherd, while he stands on the bench'}). The evaluation dataset consists of $540$ images, with one edit instruction per image. Given the generality of our method, we extend our evaluation beyond PIEBench by comparing against domain-specific baselines on tasks such as facial attribute and material editing.


\vspace{1mm}
\noindent 
\textbf{Metrics.} We evaluate all the methods on two aspects: smoothness of edit trajectories and instruction following. Smoothness is measured with the triangle deficit ($\delta_{\text{smooth}}$), which captures second-order consistency between adjacent edits; smaller values indicate smoother transitions. We use DreamSim~\citep{fu2023dreamsim} as the distance metric and report the maximum deficit per sequence. A user study confirmed that this configuration for measuring smoothness of edits aligns best with human preference (appendix Fig.\textcolor{cvprblue}{10}). We evaluate the instruction following with CLIP directional similarity (CLIP-dir.)~\citep{gal2021stylegannada} aggregated over all edit strengths. Additionally, we also present the plot between the subsequent image distance with DINO~\cite{oquab2023dinov2} vs the edit strength to evaluate the edit trajectories finely. We also present the faithfulness and fidelity tradeoff plot in appendix Fig.\textcolor{cvprblue}{7}. Full details about metrics and evaluation for identity preservation are provided in appendix.   


\subsection{Baseline Comparisons} 
\vspace{-1mm}
We compare \emph{Kontinuous Kontext} against two categories of baselines here, and with additional custom inference-based baselines in appendix Sec.\textcolor{cvprblue}{A.9}: 

\vspace{1mm}
\noindent
\textbf{i) Editing + interpolation:} We first generate a full strength edit with Flux Kontext and then produce intermediate editing using interpolation methods. We use Diffmorpher~\citep{zhang2024diffmorpher},  Freemorph~\citep{cao2025freemorph}, and a video inbetweening method WAN-2.1~\citep{wan2025} for interpolation and evaluate on PIEBench. Diffmorpher trains a LoRA on the two input images and interpolates the model weights, while Freemorph inverts the images and interpolates their attention features during denoising. Both are post-hoc heuristics applied to pretrained diffusion models, making them fragile across diverse edits. Video inbetweening methods, though explicitly trained for interpolation, perform poorly on imaginative stylization tasks since they are trained on real videos. Further, these baselines are slower as they require a cascade of models for slider-based editing. 

\begin{figure}[b]
    \vspace{-6mm}
    \includegraphics[width=0.95\linewidth]{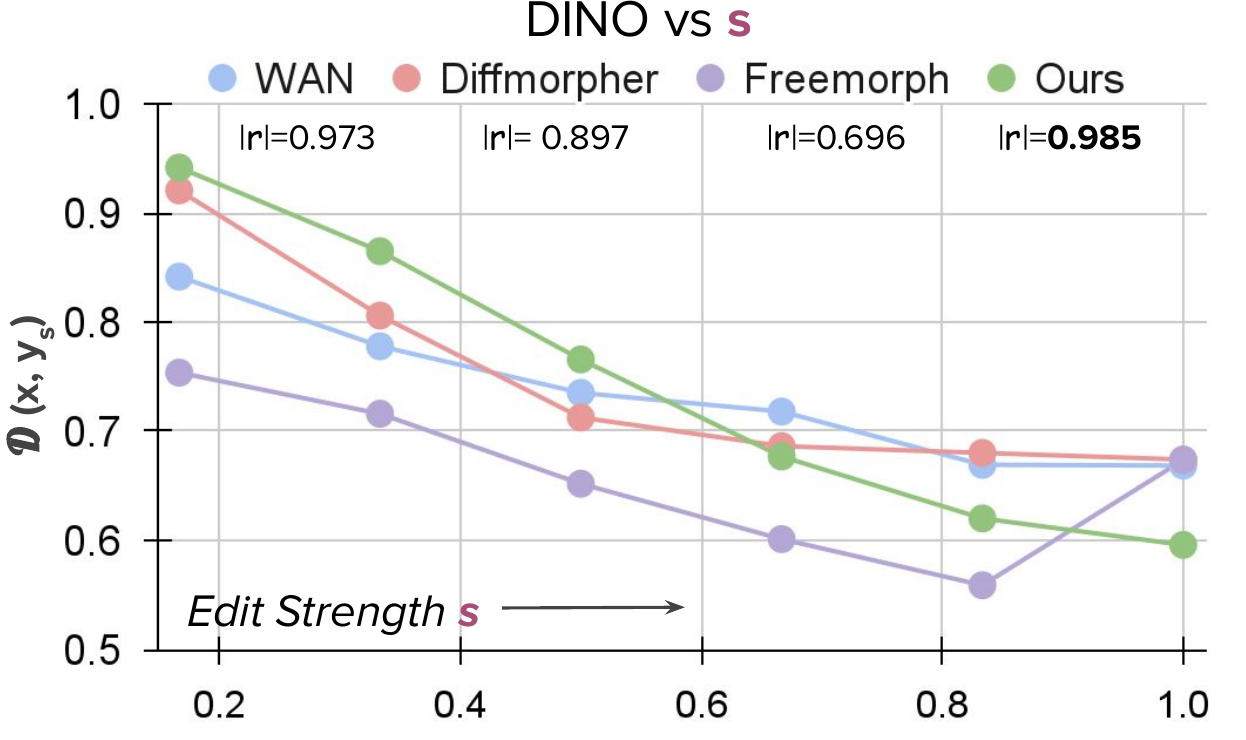}
    \vspace{-4mm}
    \caption{We plot the change in the source image as we increase the slider strength \textbf{s} with DINO. In contrast to baselines, our method gradually and monotonically changes the source image with an increase in edit strength. We quantify linearity with absolute Pearson correlation $\mathbf{|r|}$ where our method outperforms.}
    \vspace{-1mm}
    \label{fig:smoothness-plots}
\end{figure}

\begin{figure*}[t]
    \centering
    \vspace{-4mm}
    \includegraphics[width=0.95\linewidth]{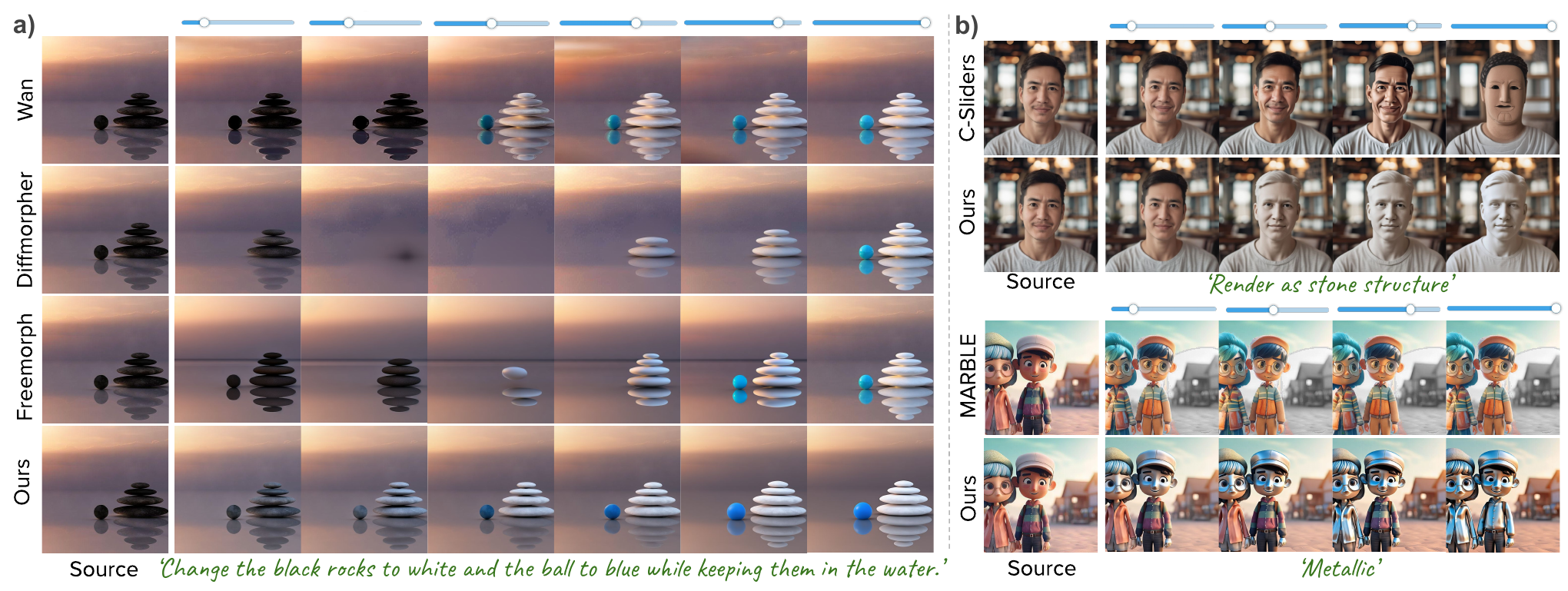}
    \vspace{-3mm}
    \caption{\textbf{Visual Comparison.} We evaluate against (a) image interpolation methods, where we first generate a full strength edit with Flux-Kontext and interpolate to obtain intermediate edits, and (b) domain-specific methods, which train separate LoRAs/Adapters for each attribute. Our generalized method achieves superior slider control with consistent image identity and smooth edit transitions.}
    \vspace{-1mm}
    \label{fig:qualitative-comparison}
\end{figure*}

\vspace{1mm}
\noindent 
\textbf{Analysis.} We present comparisons with interpolation baselines in Tab.~\ref{tab:comparison-general} and qualitative comparison in Fig.~\ref{fig:qualitative-comparison} a on a challenging composite edit. Wan inbetweening abruptly transitions the color of the objects to the target full edit as such transformations are out of distribution for the video model, which is reflected as a higher $\delta_{\text{smooth}}$ value. However, this also raises CLIP-dir, it does so only because the full edit appears prematurely at intermediate strengths. Diffmorpher and Freemorph introduce severe distortions in intermediate steps, often partially or completely removing the object, which leads to poor scores on both $\delta_{\text{smooth}}$ and CLIP-dir. Our method generates smooth transitions from the source to the final edit, gradually changing the color of the rock and ball while preserving their identity. We also evaluate the change in the source image $(x_s)$ with an increase in edit strength $\mathcal{D}(y_s, x)$ in Fig.~\ref{fig:smoothness-plots} , where $y_s$ is the edited image with strength $s$ and $\mathcal{D}$ is DINO~\cite{oquab2023dinov2} image feature extractor. Notably, our method gradually and monotonically changes the source image and has the highest linearity across baselines evaluated by Pearson correlation $|r|$ with a linear path.

\vspace{1mm}
\noindent
\textbf{ii) Domain specific methods:} Here, we compare against methods trained to control specific attributes, such as facial properties (e.g., age, smile) or material properties (e.g., transparency, metallicness). We compare with ConceptSliders~\citep{kim2025concept}, which trains a LoRA module per attribute and achieves continuous control by weight interpolation. We compare across $11$ available sliders covering facial attributes, stylization, and scene edits on a set of generated images. For material control, we compare with MARBLE~\citep{cheng2025marble}, which trains separate adapter networks similar to IP-adaptor~\cite{ye2023ipadaptertextcompatibleimage}
to achieve continuous control for each material property, such as metallicness. We evaluate MARBLE on the images from the material editing category from PIEBench, for metallicness and glow material. Details about the evaluation set are provided in appendix Sec.\textcolor{cvprblue}{A11}.


\begin{table*}[t]
\centering
\small
\setlength{\tabcolsep}{4pt}
\renewcommand{\arraystretch}{1.1}

\begin{minipage}[t]{0.325\linewidth}
\centering
\begin{tabular}{@{}lcc@{}}
\toprule
Methods & $\delta_{\text{smooth}} \downarrow$ & CLIP-Dir. $\uparrow$ \\ \midrule
Diffmorpher & 0.371 & 0.181 \\
Freemorph & \underline{0.365} & 0.189 \\
WAN-Video & 0.853 & \textbf{0.269} \\ 
Ours & \textbf{0.329} & \underline{0.241} \\ 
\bottomrule
\end{tabular}
\vspace{-2mm}
\caption{\small \textbf{Editing + Interpolation} baselines.}
\label{tab:comparison-general}
\end{minipage}%
\hfill
\begin{minipage}[t]{0.325\linewidth}
\centering
\begin{tabular}{@{}lcc@{}}
\toprule
Methods & $\delta_{\text{smooth}} \downarrow$ & CLIP-Dir. $\uparrow$ \\ \midrule
ConceptSliders & 0.143 & 0.186 \\ 
Ours & \textbf{0.098} & \textbf{0.382} \\
\noalign{\vskip -2pt}
\midrule
\noalign{\vskip -2pt}
MARBLE & 2.577 & \textbf{0.157} \\ 
Ours & \textbf{0.350} & 0.101 \\ 
\bottomrule
\end{tabular}
\vspace{-2mm}
\caption{\small \textbf{Domain specific} baselines.}
\label{tab:comparison-domain-specific}
\end{minipage}%
\hfill
\begin{minipage}[t]{0.325\linewidth}
\centering
\begin{tabular}{@{}lcc@{}}
\toprule
Methods & $\delta_{\text{smooth}} \downarrow$ & CLIP-Dir. $\uparrow$ \\ \midrule
text-space condn & 1.468 & 0.191 \\ 
w/o text projector & 1.092 & 0.141 \\ 
w/o filtering & 0.483 & 0.228 \\ 
Ours & \textbf{0.329} & \textbf{0.241} \\ 
\bottomrule
\end{tabular}
\vspace{-2mm}
\caption{\small Ablation study.}
\label{tab:ablations}
\end{minipage}
\vspace{-5mm}
\end{table*}

\vspace{1mm}
\noindent 
\textbf{Analysis}
We compare with domain-specific methods in Fig.~\ref{fig:qualitative-comparison}b and Tab.~\ref{tab:comparison-domain-specific}. In comparison to ConceptSliders (C-Sliders), our method produces smoother transitions in appearance while preserving facial structure, as reflected in lower $\delta_{\text{smooth}}$. In contrast, C-Sliders often produce weak edits (see appendix for more comparisons), resulting in lower CLIP-dir. MARBLE, trained on synthetic 3D assets for material control, struggles on complex real images and, even when successful, exhibits abrupt jumps to the final edit at lower strengths. This leads to significantly higher $\delta_{\text{smooth}}$ despite high CLIP-dir. Our method achieves smooth and consistent transitions across diverse scenarios. Importantly, unlike domain-specific approaches that require attribute-specific training, our model works out of the box for new attributes, offering a single unified solution for continuous control of diverse attributes as shown in Fig.~\ref{fig:results-ours} \& appendix Fig.\textcolor{cvprblue}{1}. Additional comparisons are in appendix Sec.\textcolor{cvprblue}{A.8}.


\subsection{Ablations} 
\label{subsec:ablate}
\vspace{-1mm}
We ablate design choices in Tab.~\ref{tab:ablations}. Conditioning by adding the slider projector output as an extra text token (\textbf{text-space condn}) is ineffective for fine-grained strength control and produces abrupt transitions, reflected in the worst $\delta_{\text{smooth}}$. Removing the pooled text embedding input from the slider projector (\textbf{w/o text projector}) leads to weaker, non-smooth edits and inferior $\delta_{\text{smooth}}$ and CLIP-dir. scores (see appendix Fig.\textcolor{cvprblue}{6}). Finally, effective data filtering that removes poor-quality and non smooth edit sequences from the dataset significantly improves both smoothness and text alignment. 


\begin{figure}[b]
    \centering
    \vspace{-5mm}
    \includegraphics[width=0.9\linewidth]{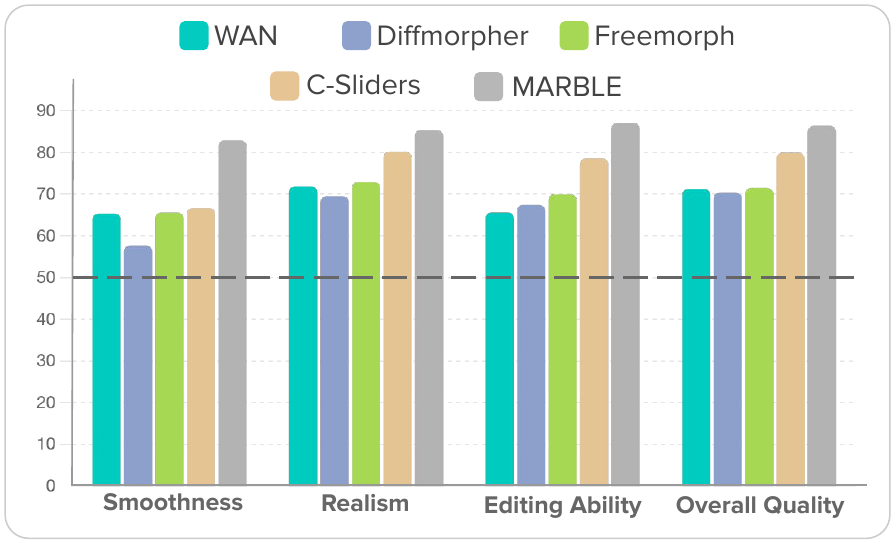}
    \vspace{-3mm}
    \caption{\small User study win rates (\%) of our method over baselines.}
    \vspace{-2mm}
\label{fig:user-study}
\end{figure}

\vspace{-1mm}
\subsection{User Study} 
\vspace{-1mm}
We conducted a user study to subjectively evaluate our method against baselines. The study followed a head-to-head comparison where for each trial, one baseline was randomly selected, and its outputs were compared with ours across four dimensions: \textit{smoothness of the edit sequence}, \textit{realism of the edits}, \textit{editing capability} with respect to the given instruction, and overall \textit{sequence quality}. For each baseline, we sampled $20$ input images, resulting in a total of $100$ images evaluated. The study involved $20$ participants, each providing judgments on the paired outputs. Figure~\ref{fig:user-study} reports the win rates of our method over the baselines. Morphing based methods often appear smooth due to continuous transitions but suffer from artifacts or missed edits. Our method consistently outperforms all baselines across all criteria, delivering both faithful edits with superior quality.

\vspace{-1mm}
\section{Discussion and Conclusions }
\vspace{-1mm}

We presented \emph{Kontinuous Kontext} that adds a continuous control dimension to instruction-driven image editing. Our method provides smooth, fine-grained control over the intensity of edits, without sacrificing the strong baseline capabilities of the underlying editing model. While highly effective for continuous edits, our approach has some limitations. As \emph{Kontinuous Kontext} is built on Flux Kontext, it inherits its weaknesses in categories like precise geometric manipulations, such as accurate object rotation or translation, where the base model itself struggles. A failure case of our method is in generating consistent extrapolating edits (appendix Fig.\textcolor{cvprblue}{4}) for large transformations. Beyond its practical utility, this work highlights that edit intensity is naturally encoded in the modulation space of instruction-driven diffusion models. By learning a lightweight projector into this space, we unlock a flexible control mechanism that generalizes across diverse edits without attribute specific training. This suggests that other forms of continuous control, such as spatial or temporal intensity fields, may be introduced in a similarly lightweight manner, opening opportunities for interactive editing tools that combine the richness of language with the precision of continuous sliders.



\noindent 
\textbf{Acknowledgements.} We thank Rinol Gal and Amil Dravid for thoroughly reviewing the manuscript. This work was supported by the Prime Minister’s Research Fellowship (PMRF) by the Government of India and the Kotak IISc AI-ML Centre. 


\appendix 

\clearpage
\setcounter{page}{1}
\maketitlesupplementary
\tableofcontents

\section{Appendix}
\begin{figure*}
    \centering
    \includegraphics[width=0.75\linewidth]{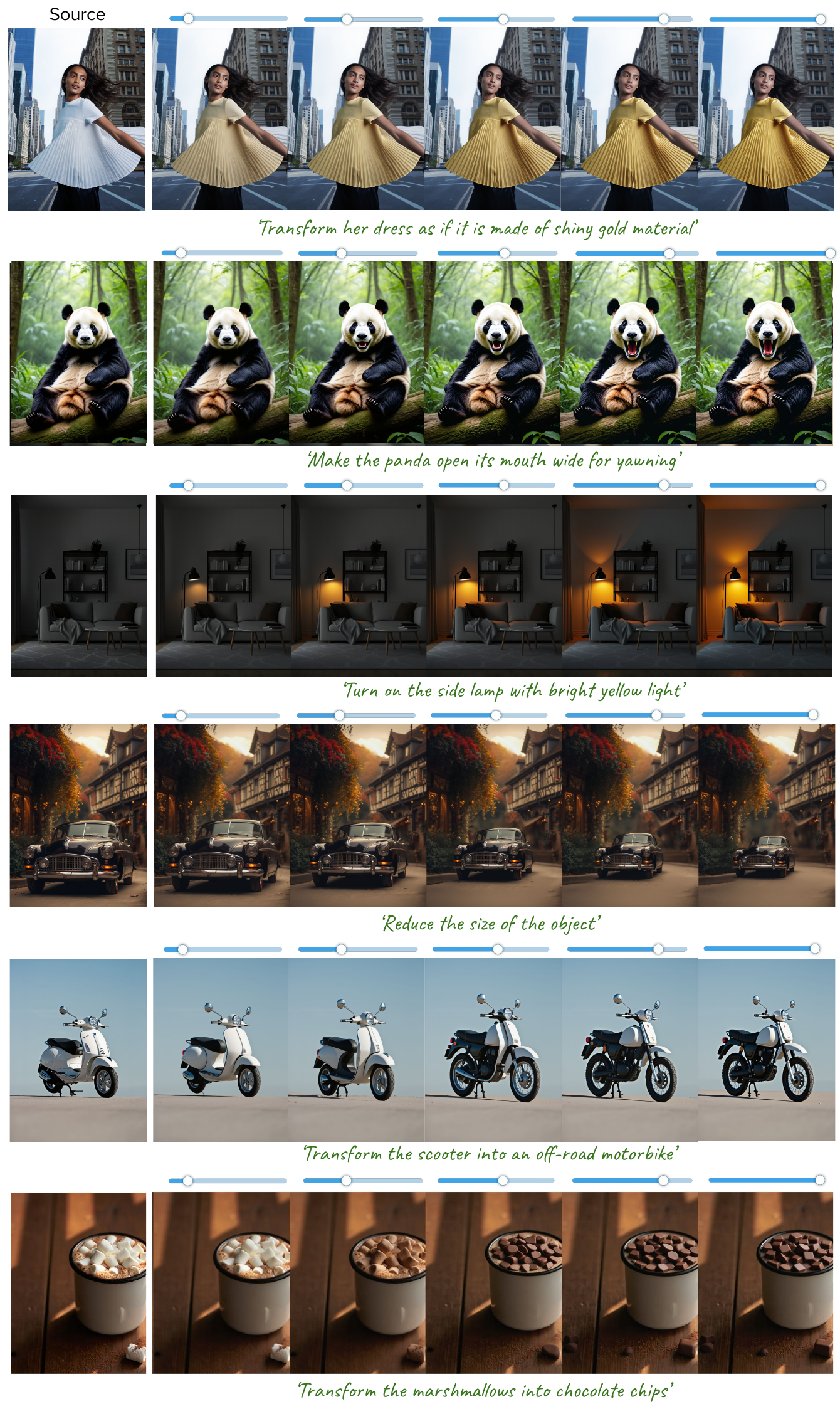}
    \caption{\emph{Kontinuous Kontext} can enable fine-grained control over the edit strength for diverse instruction-driven image editing operations.}
    \label{fig:qualitative-results-supply}
\end{figure*}

\subsection{Implementation Details.}
\label{subsec:supply-implementation}
\vspace{-2mm}
We train a slider projector along with a rank $4$ LoRA on all attention layers of the base diffusion transformer model. We train all our models at a resolution of $512 \times 512$. After filtering, our dataset consists of $~66K$ edit trajectories, along with their edit instructions. We use a threshold value of $0.15$ on the KL divergence to filter the non-smooth edit trajectories. We train all models on a $8$ NVIDIA $A100$ ($80GB$) GPU for $110,000$ iterations with an effective batch size of $8$ and a constant learning rate of $2 \times 10^{-5}$. Training takes about $72$ hours to complete. During training, we drop the slider conditioning with $10\%$ of the time. For inference, we use the default Euler scheduler from Flux Kontext and use $T=28$ inference steps for generation. The generation time is similar to the Flux Kontext model, as we only add a small MLP projector on the base model (details are in Sec.\ref{subsec:compute}).  

\subsection{Dataset Generation}
\label{subsec:supply-data-generation}
In this section, we provide the details about our dataset generation process. Our dataset generation consists of three stages:

\begin{figure*}[t] 
    \centering
    \includegraphics[width=0.90\linewidth]{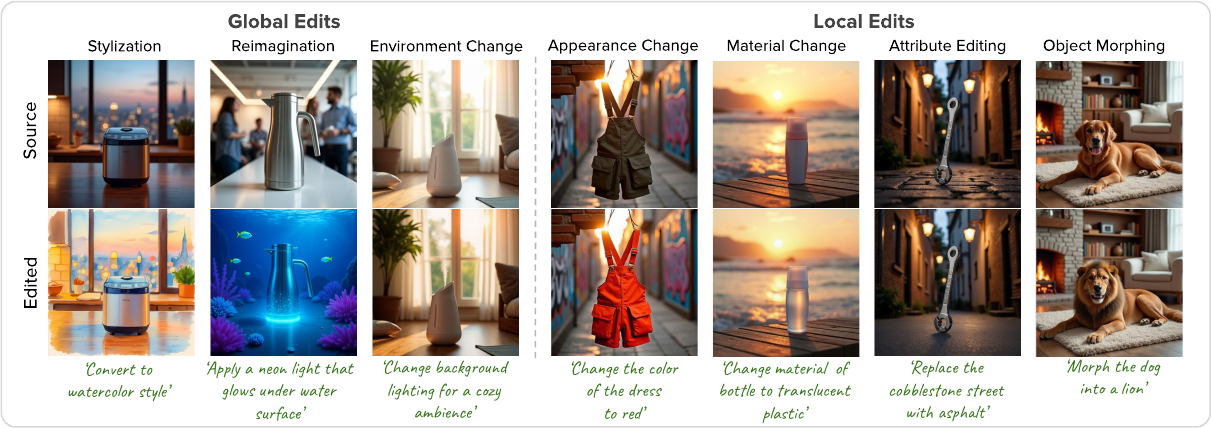}
    \caption{Samples from diverse image editing categories in our synthesized dataset spanning global edits (stylization, reimagination, and environment changes), as well as local edits (appearance changes, material changes, attribute editing, and object morphing)}
    \label{fig:data-categories}
\end{figure*} 

\vspace{2mm}
\noindent
\textbf{i) Generating Image Edit Pairs.} We use the Subject200K~\citep{tan2024ominicontrol} dataset to source our input images. This dataset consists of a variety of input objects and scenes captured in different environmental conditions. We extract $110K$ source images from this dataset. Next, we generate image-specific edit instructions for source images using Qwen-VLM~\citep{Qwen2.5-VL}. For a good diversity of our dataset, we categorize our edit categories into global edits (stylization, scene reimagination, and environment change) and local edits (material and appearance editing, attribute modification, and shape morphing). For each image in the dataset, we randomly sample one of these editing categories and query the Qwen-VLM to generate an edit instruction from that category. We pass the input image along with the system prompt to the Qwen-VLM to generate instructions specific to the image. We use the system prompt for `appearance change' edit (see \textcolor{cvprblue}{\textbf{Q-1}} on page $3$) and query the VLM to generate the edit instruction in a \textit{.json} format. We use similar system prompts for other editing categories. We sample a predefined set $50-100$ in-context examples per edit category and randomly sample $4$ examples and combine them with the system prompt to generate rich prompts for generating diverse editing instructions. 
We present samples of in-context examples used in \textcolor{cvprblue}{\textbf{Q-2}} and \textcolor{cvprblue}{\textbf{Q-3}} on pages $3$ and $4$. 

\begin{figure*}
\begin{tcolorbox}[colback=gray!10, colframe=gray!40!black, 
                  title=(Q-1) System prompt for generating edit instructions, 
                  fonttitle=\bfseries, 
                  boxrule=0.5pt, arc=2mm, left=2mm, right=2mm]
System Prompt: \texttt{You are a professional image editor. Generate an original, diverse, and detailed local appearance change instruction for the given object in the image. Create a unique instruction different in wording and content from the examples.}

\texttt{Examples:} \{examples\}

\texttt{Output ONLY a valid JSON object with EXACT keys 
\texttt{"category"} and \texttt{"instruction"}. 
No additional text or explanation.}

\texttt{Example output: 
\{"category": "Appearance\_Change", "instruction": 
"Modify the fabric of the couch to a rich burgundy velvet 
with gentle sheen."\}  
DO NOT include trailing commas or escape characters.}
\end{tcolorbox}
\end{figure*}

\noindent 

\begin{figure*}[t]
\begin{tcolorbox}[colback=gray!10, colframe=gray!40!black,
                  title=(Q-2) In context example for local edits,
                  fonttitle=\bfseries,
                  boxrule=0.5pt, arc=2mm, left=2mm, right=2mm]
\ttfamily
\textbf{Appearance\_change} \\
    examples = [ "Transform the chair into plush candy-colored marshmallow material with soft reflections", \\
    \ \ "Make the bicycle frame appear as flowing liquid metal with dynamic highlights", \\
    \ \ "Turn the lampshade into glowing crystalline material with internal refracted light"] \\[6pt]

\textbf{Material\_change} \\
    examples = [ "Replace the chair’s wooden legs with polished chrome metal, emphasizing its reflective specularity", \\
    \ \ "Make the tabletop appear carved from dark mahogany wood with visible grain and a semi-matte roughness", \\
    \ \ "Transform the bag’s fabric into smooth black leather with glossy highlights and subtle texture"] \\[6pt]

\textbf{Attribute\_change} \\
    examples = ["Open the laptop lid halfway to reveal the keyboard", \\
    \ \ "Rotate the handlebar of the bicycle by 45 degrees", \\
    \ \ "Raise the adjustable lamp arm to maximum height"] \\[6pt]
    
\textbf{Intra\_object\_morph} \\
    examples = ["Morph a teapot into a lantern while keeping the spout as a decorative handle", \\
    \ \ "Transform a bicycle into a motorbike with parts composed naturally", \\
    \ \ "Morph a chair into a bench while preserving the backrest shape"] \\[6pt]
\end{tcolorbox}
\end{figure*}

\begin{figure*}[t]
\begin{tcolorbox}[colback=gray!10, colframe=gray!40!black,
                  title=(Q-3) In context example for global edits,
                  fonttitle=\bfseries,
                  boxrule=0.5pt, arc=2mm, left=2mm, right=2mm]
\ttfamily
\textbf{Stylization} \\
    examples = [ "Render the scene in Studio Ghibli style with dreamy backgrounds and soft pastel hues", \\
    \ \ "Transform the image into Pixar-style 3D animation with vibrant colors and cinematic lighting", \\
    \ \ "Stylize the composition as a Van Gogh oil painting with thick impasto brush strokes"] \\[6pt]

\textbf{Environment\_change} \\
    examples = ["Blanket the entire landscape with fresh, thick snow, covering trees and rooftops with crystalline frost", \\
    \ \ "Transform the scene into a harsh winter blizzard with swirling snow and reduced visibility", \\
    \ \ "Age the entire scene to look like a weathered medieval village with cracked stone walls"] \\[6pt]

\textbf{Scene\_reimagination} \\
    examples = ["Place the entire village on a massive turtle’s back slowly moving through the ocean", \\
    \ \ "Transform the bustling marketplace into a floating bazaar carried by hot air balloons", \\
    \ \ "Reimagine the city skyline as colossal crystal formations reflecting rainbow light"]
\end{tcolorbox} 
\end{figure*}

\begin{figure*}[t]
    \centering
    \includegraphics[width=\linewidth]{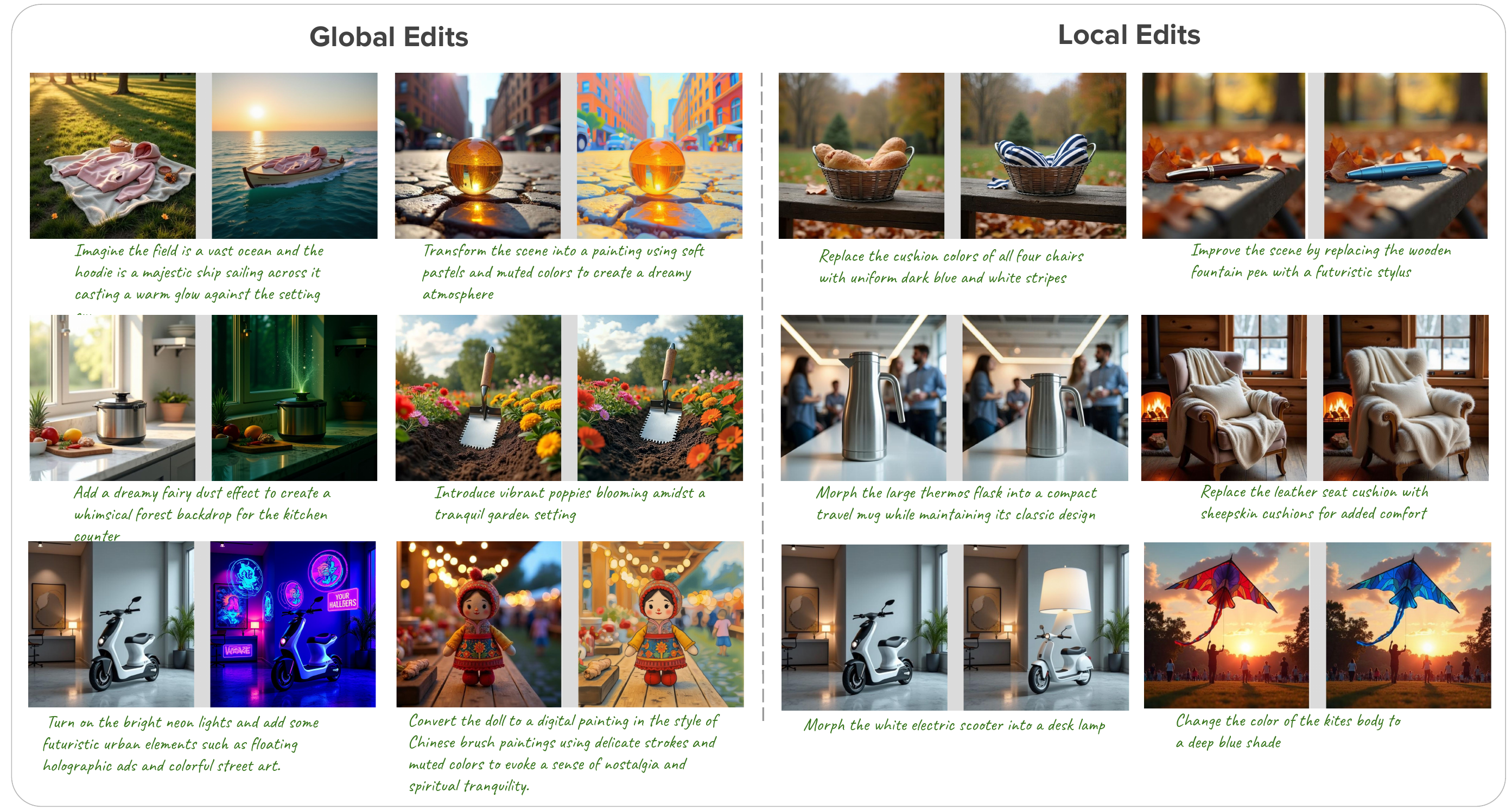}
    \caption{Samples for generated edit instructions and the generated edits from Flux Kontext}
    \label{fig:edit-samples-supply}
\end{figure*}

\vspace{2mm}
\noindent 
\textbf{Generating image edits.} We use the source images and obtain editing instructions to generate edited versions of the source image using Flux Kontext~\citep{batifol2025flux}. Flux-kontext, being a generalist editing model, can generate high-quality edits for the source images given text prompts. However, in some cases, it does not perform the edit and outputs the same input image. We filter such cases in our filtering stage, discussed next. We show some examples of the edits and instructions in Fig.~\ref{fig:edit-samples-supply}. 

\begin{figure*}[h]
    \centering
    \includegraphics[width=0.9\linewidth]{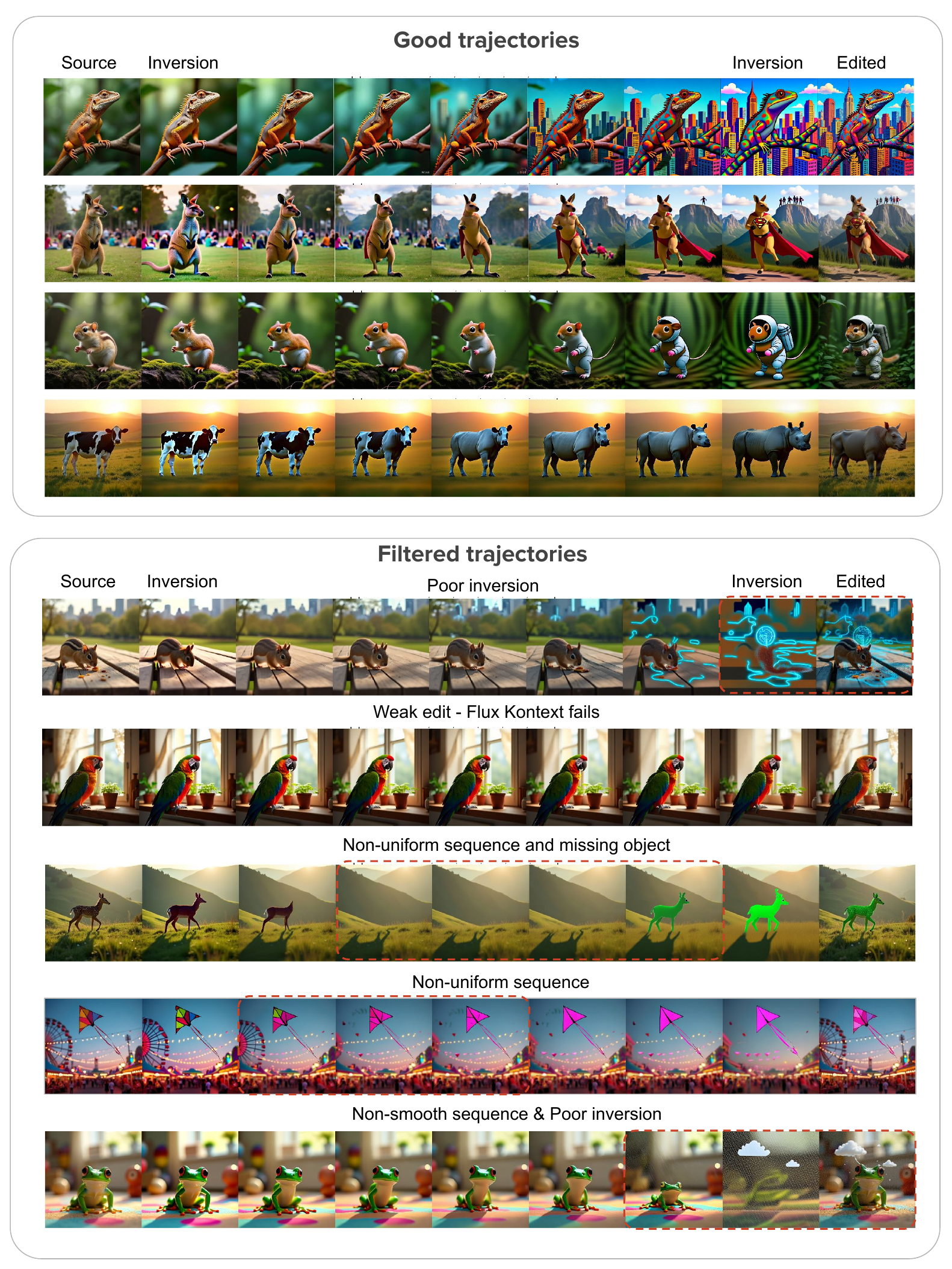}
    \caption{Samples trajectories from our synthesized dataset}
    \label{fig:data-filtering-supply}
\end{figure*}

\vspace{2mm}
\noindent
\textbf{ii) Generating intermediate edits with Image morphing}. Given the source and edited image, we use Freemorph~\citep{cao2025freemorph} - a training-free Diffusion-based image morphing approach to generate interpolations. These interpolations will be used as ground truths for the edits with intermediate strengths. Freemorph requires an input caption for the two images to be interpolated. To this end, we use LLaVA~\citep{llava} to generate captions as suggested by their paper. Freemorph first inverts the two images and then interpolates the attention features during denoising. This requires a full denoising process to generate one morph image. In practice, we generate $N=5$ intermediate morphs between the source and the edited image. We use the official code provided by the authors that is built on StableDiffusion-$2.1$~\citep{rombach2022high} and use the DDIM scheduler for generation with $T=50$ steps. All the interpolations were generated at a native resolution $768 \times 768$ of SD-$2.1$ base model. Though Freemorph generates smooth interpolations in most of the cases, it may contain errors in some, that we filter out with our extensive filtering pipeline

\begin{figure*}[h]
    \centering
    \includegraphics[width=\linewidth]{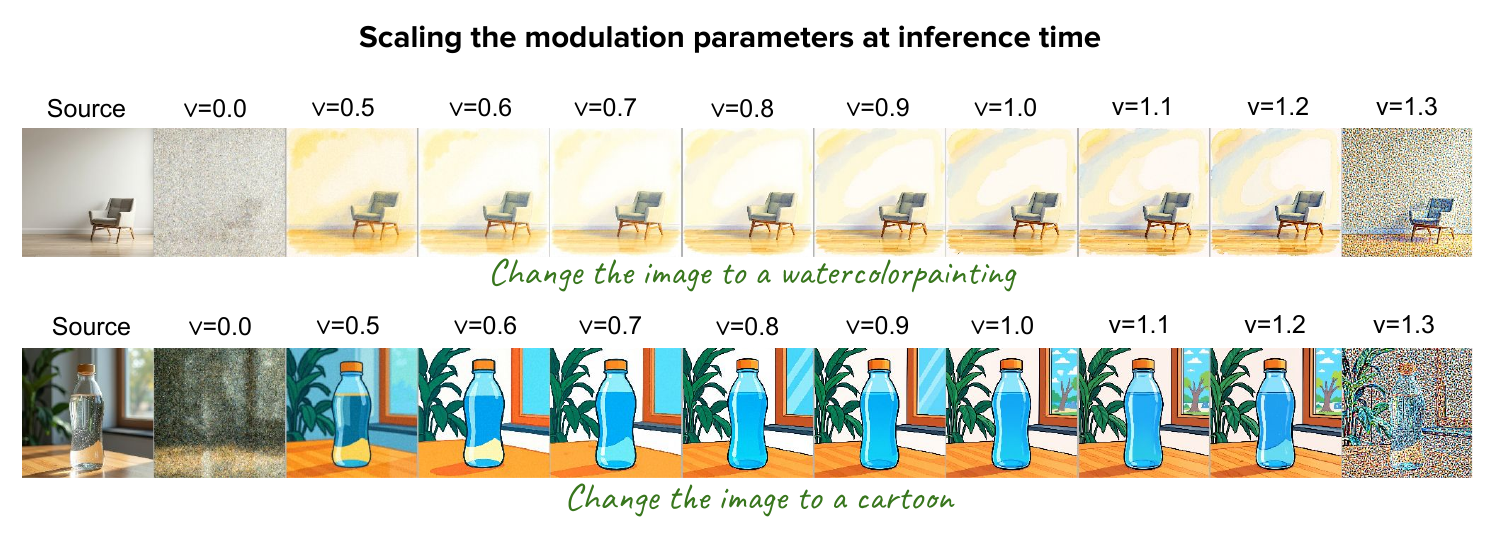}
    \caption{\textbf{Inference time control in modulation space.} We conducted a simple experiment by scaling the text modulation parameters with values of $v \in (0.5, 1.3)$ to generate multiple edits. While these edits varied across different scales, the variations did not consistently correlate with the intended edit strength. This highlights the need for a dedicated learning module that can translate such variations into user-interpretable strength control by accurately manipulating the modulation parameters.}
    \label{fig:supply-modulation-toy}
\end{figure*}

\noindent
\textbf{iii) Data Filtering.} As our data generation process involves the use of several generative models, it may lead to errors in some of the generated data samples. We filter such cases with an automated data filtering logic. Specifically, we remove samples that have - \textbf{\textit{a)}} inversion errors, \textbf{\textit{b)}} non-smooth trajectories, or \textbf{\textit{c)}} weak edits. Some examples are shown in Fig.~\ref{fig:data-filtering-supply}.

\vspace{2mm}
\noindent 
\textbf{a) Poor inversion.} We use Freemorph for generating image edit sequences that involve diffusion inversion of the input images. The inversion of the source and the edited image can be error-prone, resulting in an inaccurate training signal. To address this, we compute the LPIPS distance between the original image and its inversion and remove the sample if the LPIPS is above a threshold, indicating poor inversion. Some examples are shown in the first and last row in Fig.~\ref{fig:data-filtering-supply}.  

\vspace{2mm}
\noindent 
\textbf{b) Non-smooth trajectories.} We also filter out samples where the edit sequence has non-uniform transitions, i.e., change between the two adjacent images is not consistent across the sequence (see rows $3$,$4$,$5$ in Fig.~\ref{fig:data-filtering-supply}). For a good training sample $(x,e,s,y_s)$, the extent of change between the source $x$ and edit $y_s$ should scale with the edit strength $s$. Equivalently, the distance between adjacent images in the sequence should remain consistent. We define the sequence of deltas as
$D = \{d_{0,1}, d_{1,2}, \ldots, d_{N-1,N}\}$, where $d_{i,i+1}$ is the distance between image $y_i$ and $y_{i+1}$ and measures its uniformity via the KL-divergence from a discrete uniform distribution. In an ideal edit sequence, this sequence of image distances should follow a uniform distribution. We use a KL threshold value of $0.15$ to filter such cases with non-smooth trajectories. 

\vspace{2mm}
\noindent 
\textbf{c) Weak edits.} As we use Flux Kontext to generate the edited version of the source image, we filter out cases where Flux Kontext is not able to perform the edit (see row $2$ in Fig.~\ref{fig:data-filtering-supply}). In such cases, Kontext replicates the source image as output without any meaningful change. We filter such cases by computing the LPIPS between the source image and the output image and thresholding on LPIPS value.  

\vspace{2mm}
\noindent 
\textbf{Deciding filtering thresholds.} We select these thresholds by manually inspecting a small set of edit sequences ($\approx 100$) per edit category. Additionally, one can use a VLM to automatically decide on these thresholds if data needs to be added from new other sources. 

\subsection{Model Architecture}
\label{subsec:suppy-model-arch}
Our projector is a $4$ layer MLP with dimensions $1536 \rightarrow 6144 \rightarrow 6144 \rightarrow 6144 \rightarrow 6144$. The output dimension of $D=6144$ is divided into two chunks, each of $3072$, representing offsets for modulation parameters - $\Delta y_{scale}$ and $\Delta y_{shift}$. The $1536$ dimensional input to the model consists of an embedded scale value $s$ of dimension $768$ and a pooled CLIP text embedding of dimensions $768$. We first apply sinusoidal positional encoding to $s$ to bring it to $128$ dimensions, followed by a linear layer to transform it to a similar dimension of $768$. The CLIP embedding and the encoded scale embeddings are concatenated and passed as a single input to the projector network. As our slider projector is lightweight and requires negligible computation and inference overhead over the base Flux Kontext model.

\subsection{Additional ablations and evaluations.}

\subsubsection{Inference-time control in modulation space}
\label{subsec:supply-motivation-expm}
We perform a simple experiment to analyse the effect of modulation parameters on the edited images. We scale the modulation parameters with $v=(0.5,1.3)$ for the text token and visualize the generated edit image in Fig.~\ref{fig:supply-modulation-toy}. Though the generated edits are diverse for different scale values, the scaling value $v$ does not directly correlate with the strength of the edit. This raises a need of learning a calibrated mapper like our slider projector, which can expose accurate strength control by properly manipulating the modulation parameters.

\subsubsection{Evaluation of identity preservation} 
\begin{figure*}
    \centering
    \includegraphics[width=\linewidth]{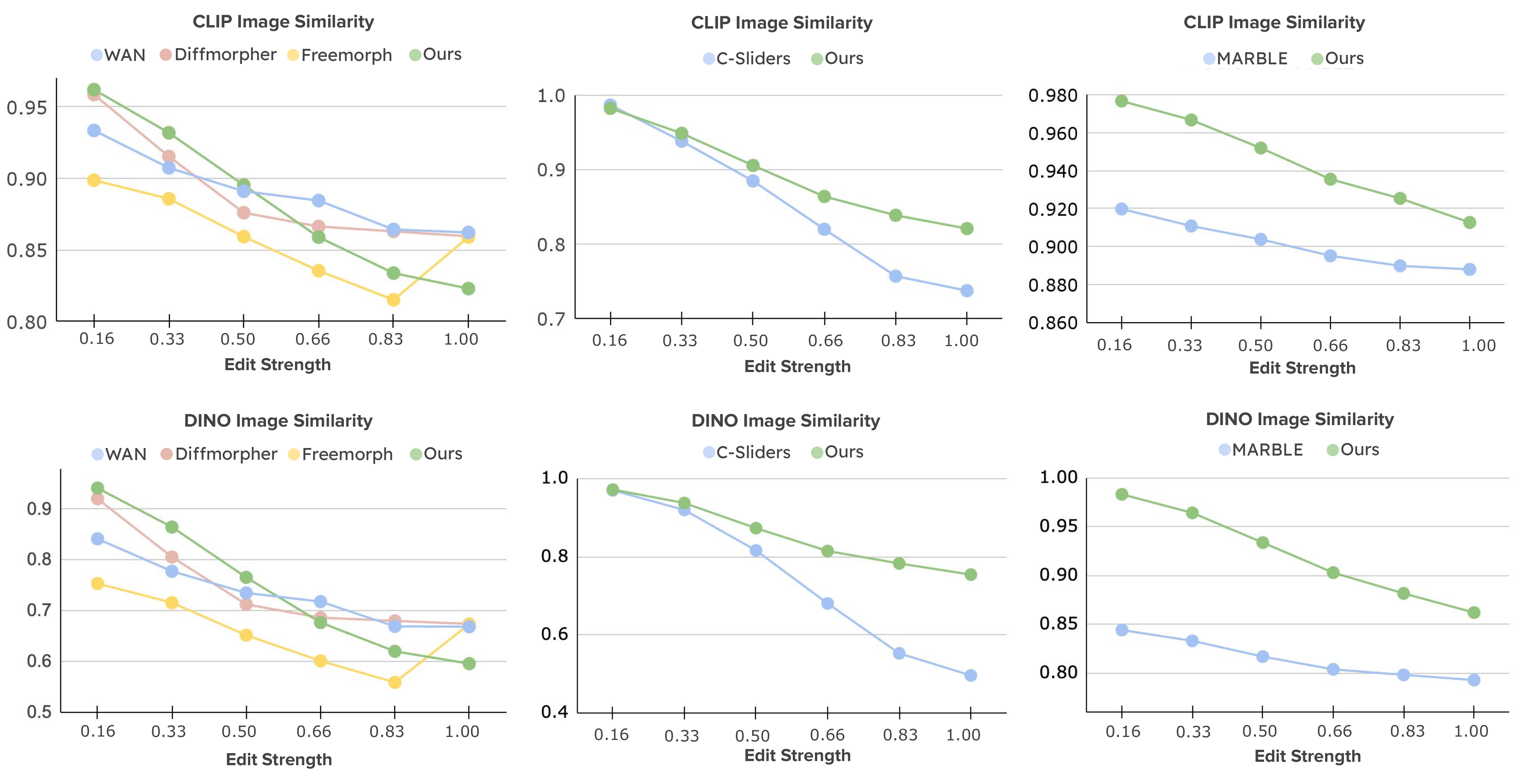}
    \vspace{-2mm}
    \caption{\textbf{Identity preservation comparison.}
    We evaluate the identity preservation of the source image during editing by computing image similarity with CLIP and Dino features. We plot the similarity scores across each edit strength to analyse the trend. Ideally, as we increase the edit strength, the image similarity should decay linearly. As compared to the baseline methods, our method achieves smooth decay trend for comparison against all the baselines. Notably, for MARBLE the image similarity is small even for the first edit, indicating its inferiority in identity preservation. For Concept-Sliders though the trend is good, it significantly changes the subject identity for higher edit strengths, resulting in poor image similarity. Our method achieves the desired trend while keeping the identity intact, even for the strongest edits, with a strength close to full edit.}
    \label{fig:supply-plot-main-metrics}
\end{figure*}

We quantify the image identity preservation by computing the CLIP-Image similarity and DINO-Image similarity between the source image and the edited image across different edit strengths. We present a plot of the image similarity value across the edit strengths in Fig.~\ref{fig:supply-plot-main-metrics}. In an ideal case, the image similarity should decay linearly with the increase in edit strength $s$. Our method achieves the desired trend while still keeping the images with the strongest edits ($s=1$) close to the source image. In comparison, the baselines either do not follow the linear trend or change the identity substantially when strong edits are performed (MARBLE, Concept Sliders).

\subsubsection{Faithfulness and quality tradeoff.}
\begin{figure}
    \centering
    \includegraphics[width=\linewidth]{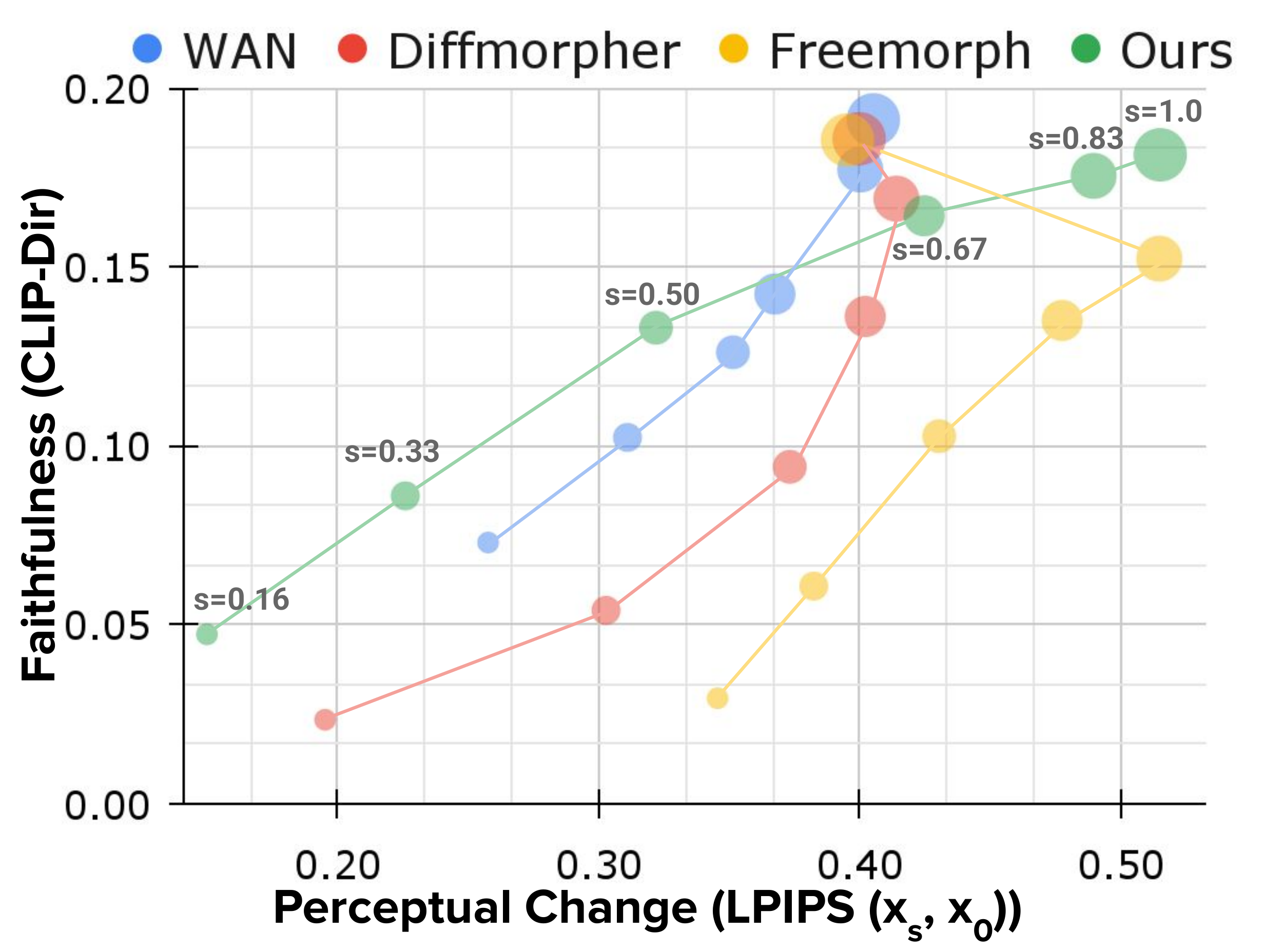}
    \caption{We plot the faithfulness of the edits along with the perceptual change in the source image. Our method achieves a monotonic and smooth transitions for both the metrics indicating superior tradeoff.}
    \label{fig:faithfulness-smoothness-tradeoff}
\end{figure}

We plot the Pareto front between faithfulness (CLIP-dir) and identity preservation (LPIPS($x_s$,$x_0$)) across all edit strengths in  Fig.~\ref{fig:faithfulness-smoothness-tradeoff}. Our method exhibits monotonic and smooth transition for both metrics across the full range of $s$. In contrast, baselines quickly diverge (higher LPIPS) from the source even for small edit strengths ($0<s<0.50$) or are not monotonic. Notably, at intermediate strength (e.g., $s=0.50$), we achieve significantly higher faithfulness with lesser identity change than baselines.

\begin{figure*}[t]
    \centering
    \includegraphics[width=0.90\linewidth]{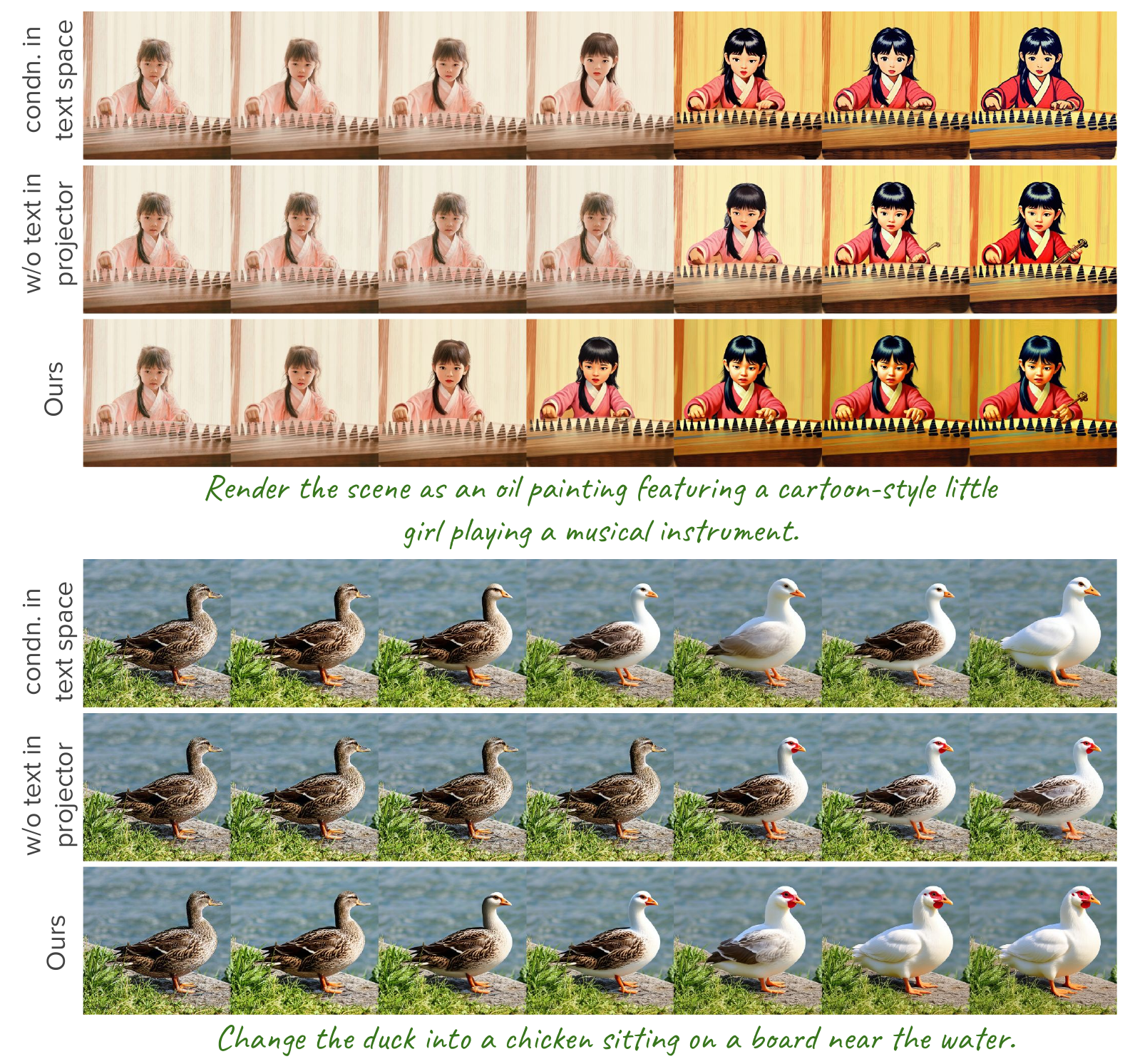}
    \caption{\textbf{Ablation over model architecture.} Injecting the slider projector in the text space results in an abrupt transition in the first example and is non-monotonic in the second example. Further, using a slider projector in the modulation space without the pooled text embedding input results in sudden transitions. Our projector takes the text embedding as input and makes appropriate adjustments to the modulation parameters for smooth transitions in the edit sequences.}
    \label{fig:supply-ablate}
\end{figure*}

\subsection{Ablation study}
\label{subsec:supply-ablate-figure}
We present an ablation study in Fig.~\ref{fig:supply-ablate} for different architecture choices. Adding the output of the slider projector in the text embedding space leads to edit transitions with abrupt jumps. Similarly, without adding the pooled text embedding in the projector leads to non-smooth edit trajectories. Our design of injecting the slider control in the modulation space and making the projector adapt to the edit instruction embedding results in smooth trajectories, enabling fine-grained control to the user.

\subsection{Evaluation Metrics}
\label{subsec:supply-metric}
\subsubsection{Smoothness of the edit sequence}
We consider first and second-order smoothness of an edit trajectory for quantitative evaluation. For a given source image $x$ and edit instruction, we generate a sequence of $N$ edited images 
$\{y_{s_1}, y_{s_2}, \ldots, y_{s_N}\}$, and include the source image as the initial element 
$y_{s_0} = x$, yielding a sequence of $N{+}1$ images. 
We use an image distance metric $d(\cdot,\cdot)$ to compare the images and explore a semantic metric (Dreamsim~\citep{fu2023dreamsim}) and LPIPS.

\begin{figure*}[t]
    \centering
    \includegraphics[width=0.90\linewidth]{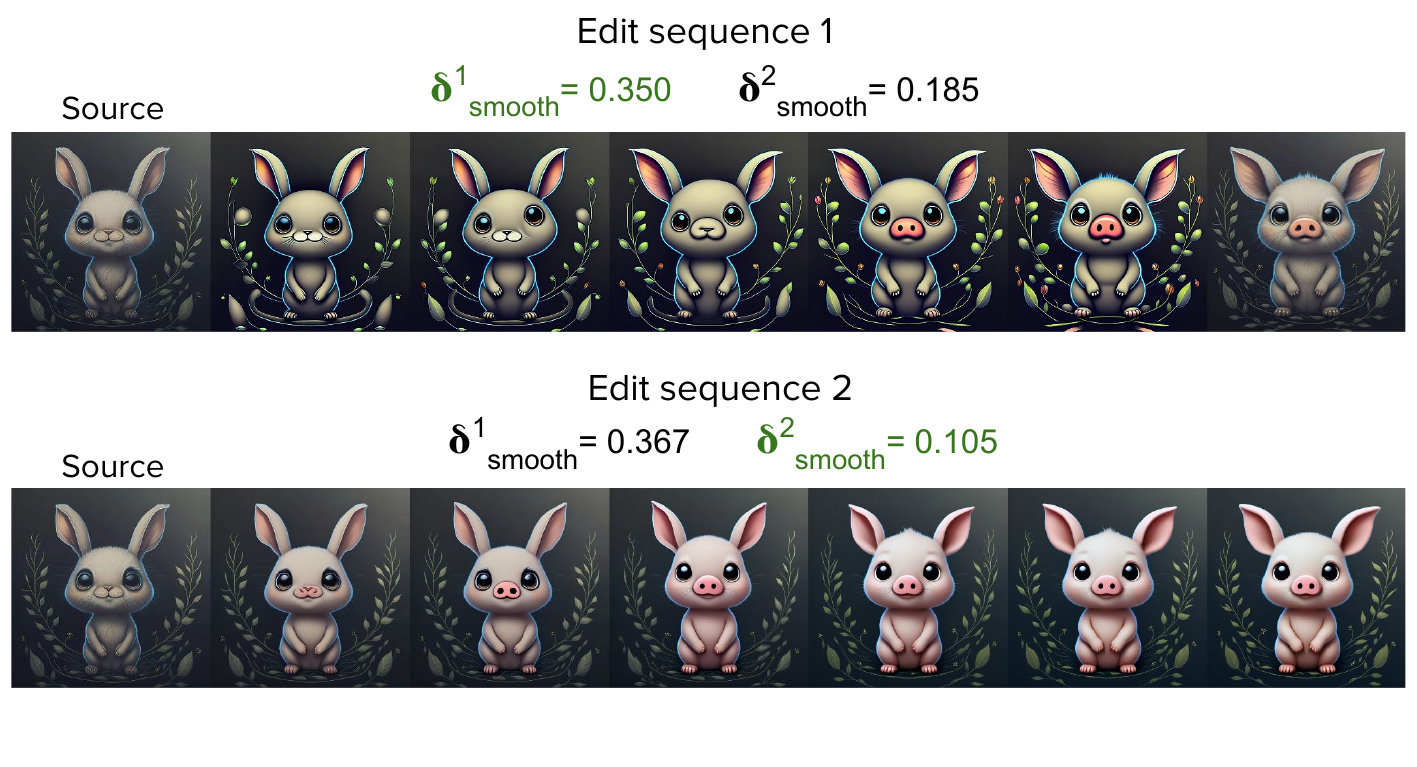}
    \caption{\textbf{Qualitative interpretation for first order and second order smoothness.} For slider-based image editing, second-order smoothness is more important than first-order smoothness, as it captures the local consistency needed for gradual, nuanced changes with slider.}
    \label{fig:smoothness-metric}
\end{figure*}

\vspace{2mm}
\noindent
\textbf{First-order smoothness.}  
We define adjacent distances between the images in the sequence as:
\[
d_i = d(y_{s_i}, y_{s_{i+1}}), \quad i = 0,\ldots,N{-}1,
\]
and compute the cumulative path length
\[
L = \sum_{i=0}^{N-1} d_i.
\]
The first-order smoothness is then computed as:
\[
\delta^1 = \max_i \frac{d_i}{L},
\]
which captures the largest normalized jump in the generated edit trajectory.  

\vspace{2mm}
\noindent 
\textbf{Second-order smoothness.}  
For local consistency, we compute the triangle deficit given by

\begin{align}
\Delta_i = d(y_{s_i}, y_{s_{i+1}}) &+ d(y_{s_{i+1}}, y_{s_{i+2}}) - d(y_{s_i}, y_{s_{i+2}}) \notag \\
i &= 0,\ldots,N{-}2. \notag
\end{align}

Each deficit is normalized by the direct distance between the endpoints:
\[
\tilde{\Delta}_i = \frac{\Delta_i}{d(y_{s_i}, y_{s_{i+2}})}.
\]
The second-order smoothness is then computed as:
\[
\delta^2 = \max_i \tilde{\Delta}_i,
\]
where smaller values indicate smoother local transitions.

\begin{figure}[h]
    \centering
    \includegraphics[width=\linewidth]{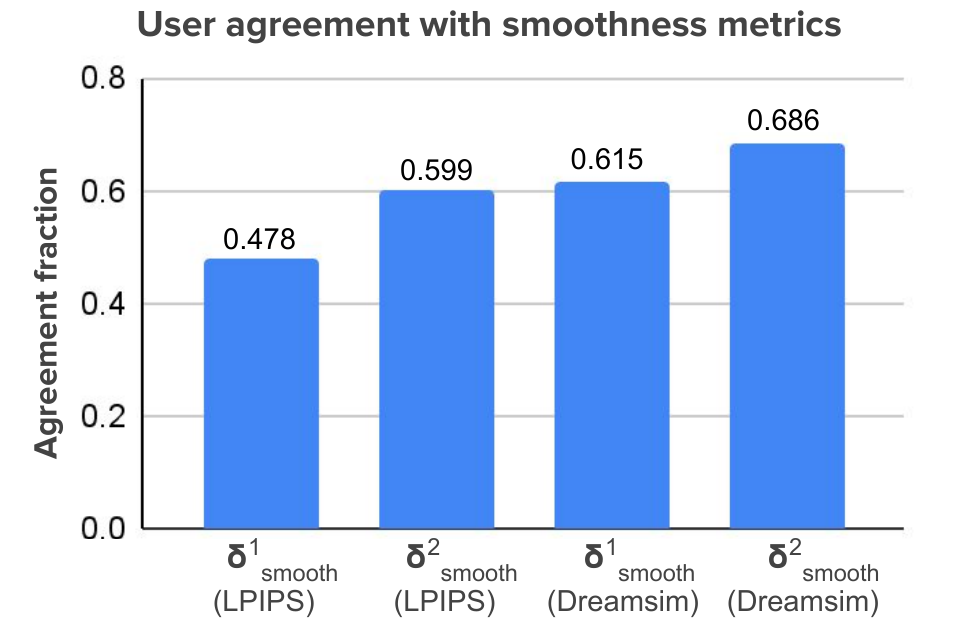}
    \caption{\small We perform a user study where we computed the alignment of the users' scores given for the smoothness of the sequence with the different variations of smoothness metrics. We found $\delta^{(2)}_{smooth}$ aligns well with the user preferences for smoothness, indicating that it is a good metric to measure the smoothness.}
\label{fig:user-alignment-study}
\end{figure}

\vspace{2mm}
\noindent
\textbf{Analysis.} We conducted a user study to evaluate how well smoothness metrics align with human preferences. Participants were shown pairs of edit sequences and asked which of the two images have smoother transitions. The study included $20$ volunteers and $40$ sequence pairs. For each sequence, we computed first- and second-order smoothness using two distance functions: LPIPS~\citep{zhang2018unreasonable} and DreamSim~\citep{fu2023dreamsim}. We then measured agreement between user choices and each of the four metric configurations (Fig.~\ref{fig:user-alignment-study}). Results show that $\delta^2$ (DreamSim) aligns best with user preferences, as it captures fine-grained semantic changes reflected in slider adjustments. While first-order smoothness prevents abrupt jumps, second-order smoothness ensures consistency in the rate of change, producing natural and continuous transitions that match user expectations for image editing. Fig.~\ref{fig:smoothness-metric} illustrates this: although Sequence $1$ has better first-order smoothness (lower $\delta^1$), Sequence $2$ is semantically smoother, captured by a lower $\delta^2_{\text{smooth}}$. From these findings, we define the smoothness metric as:
\[
\Large
\delta_{smooth} = \delta^2 (Dreamsim)
\]

\subsubsection{Instruction following with CLIP directional \\  similarity} 
For a given input image $x$, and edit instruction $e$, we edit the image with uniformly sampled edit strengths $\{s_i = i/N | i = 1, ..., N\}$ to obtain the edited image sequence $\{y_i | i = 1, ..., N\}$. We compute the CLIP-direction similarity~\citep{gal2022stylegan} for each of the edits at each strength as: 
\[
d_i = d_{clip-sim}(y_{s_i}, x, e), \\ \quad i = 1, ..., N
\]

\noindent
and report the aggregated normalized CLIP-sim as:
\[
D_{clip-dir} = \frac{\sum_{i=0}^{N} (d_i/s_i)}{N}
\]
adjusting the directional similarity based on the edit strength.

\begin{figure*}
    \centering
    \includegraphics[width=0.85\linewidth]{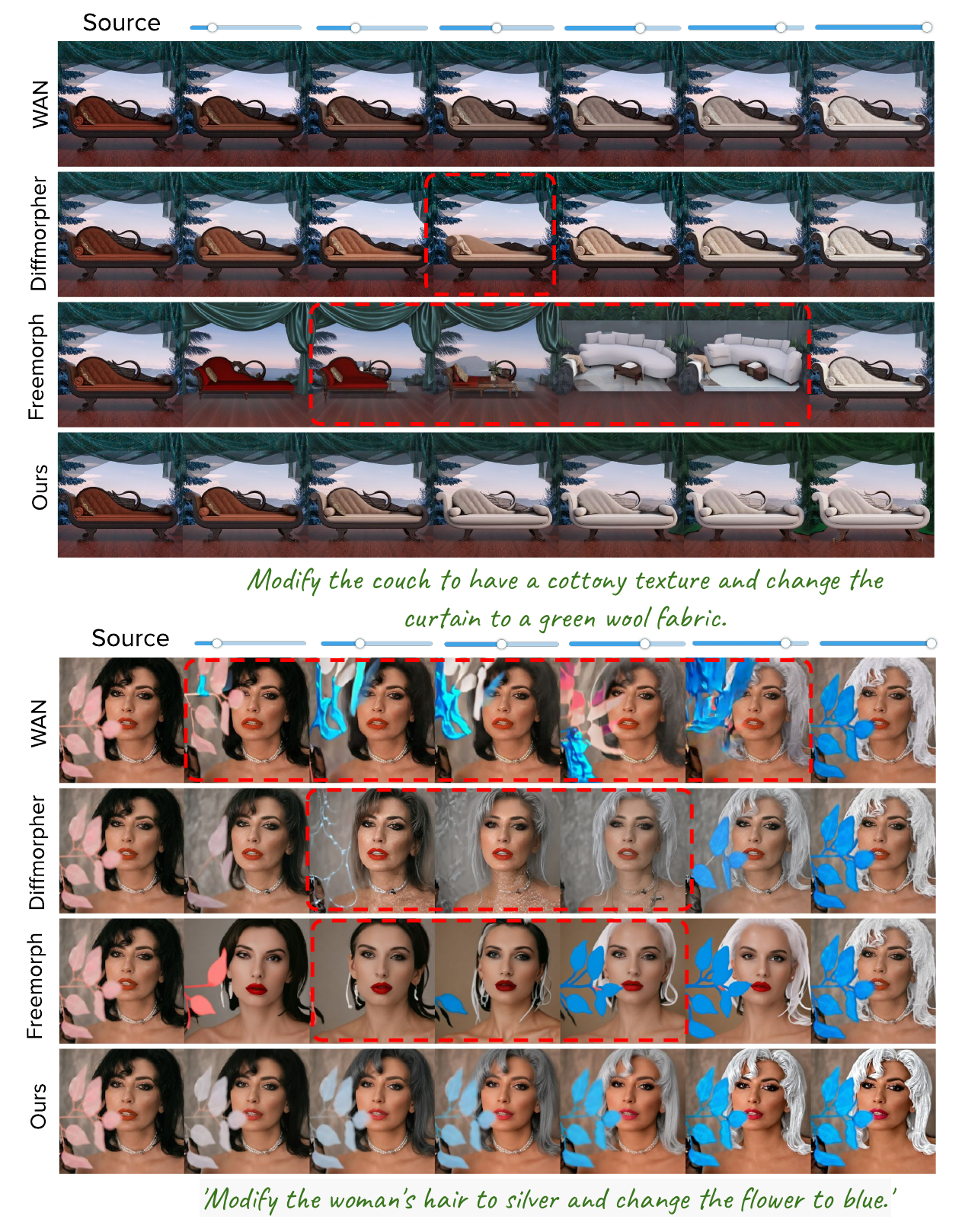}
    \vspace{-4mm}
    \caption{\textbf{Comparison with interpolation baselines.} Morphing-based methods generate smooth transitions; however, they often introduce artifacts in the intermediate images or omit details such as leaves. Similarly, the video inbetweening model WAN produces strong artifacts in intermediate frames, as these appearance transitions are out of the domain for an inbetweening model trained only on real videos.}
    \label{fig:supply-baseline-1}
\end{figure*}

\subsection{Qualitative Comparison}
\begin{figure*}
    \centering
    \includegraphics[width=0.85\linewidth]{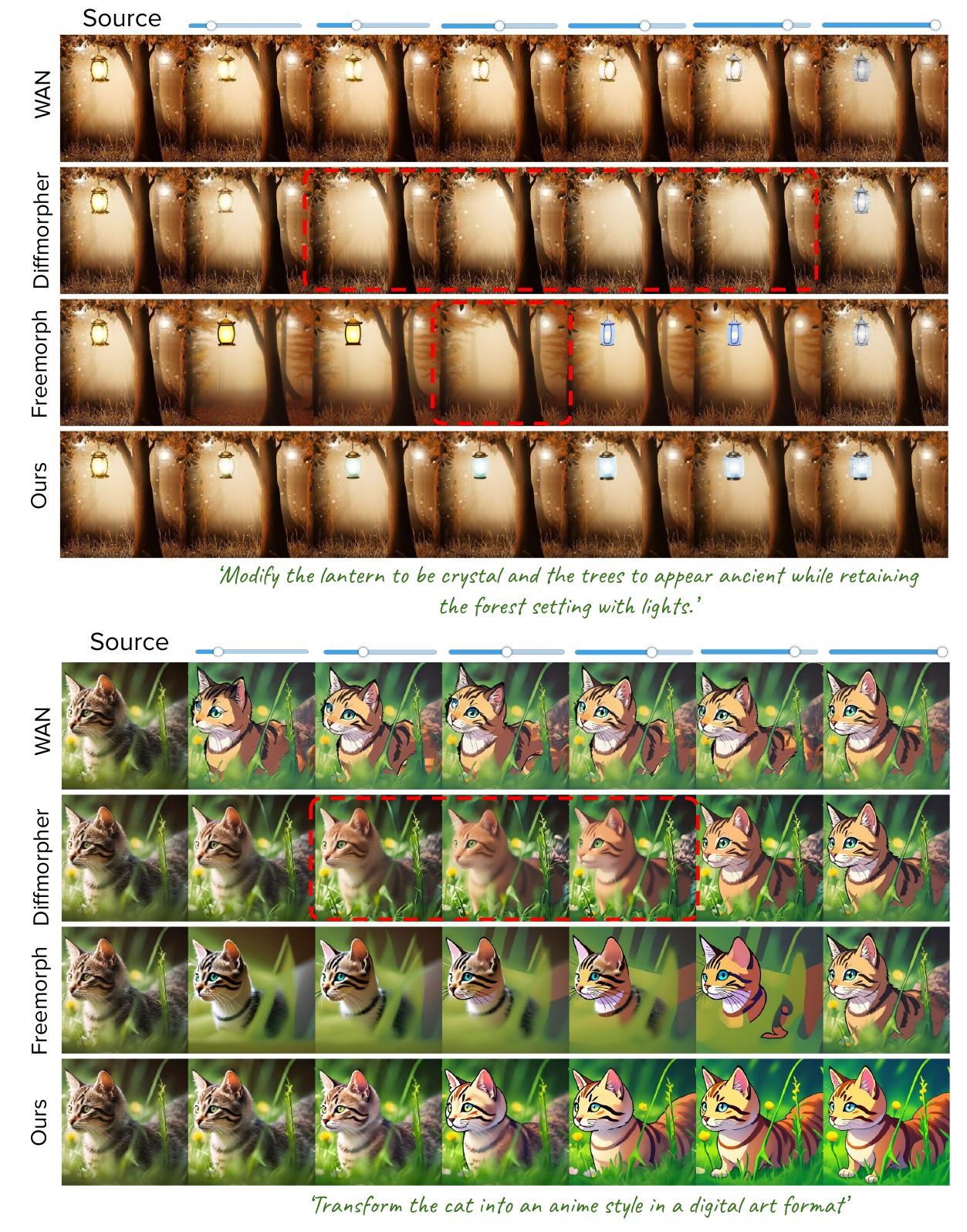}
    \caption{\textbf{Comparison with interpolation baselines.} DiffMorpher and FreeMorph remove objects in the intermediate edits of the first examples. DiffMorpher produces blurred outputs even for simple stylization transitions. The WAN inbetweening model generates transitions with abrupt jumps in both examples. In contrast, our method produces smooth transitions while preserving image identity.}
    \label{fig:supply-baseline-2}
\end{figure*}

\begin{figure*}
    \centering
    \includegraphics[width=0.85\linewidth]{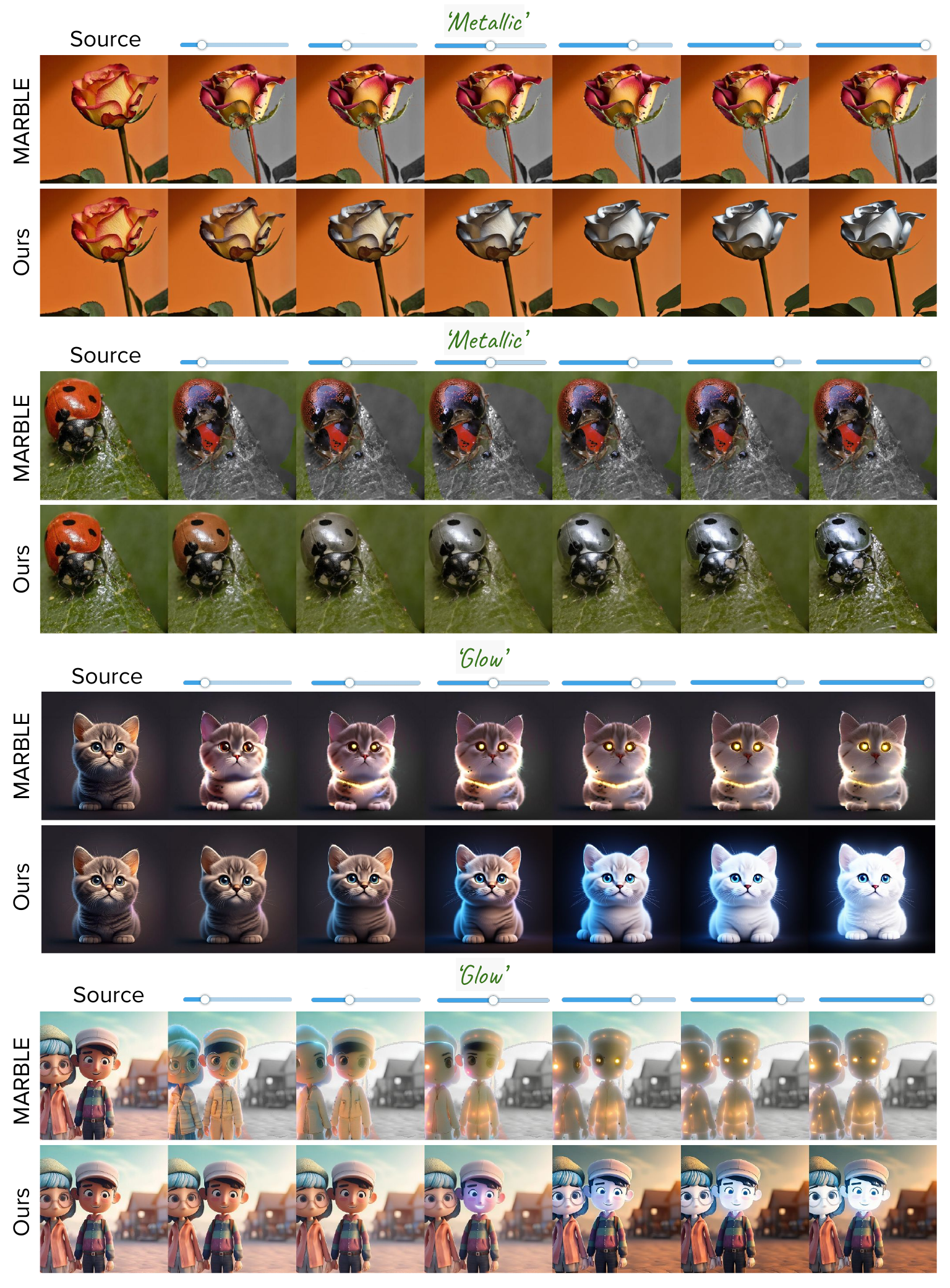}
    \caption{Comparison with MARBLE for material control}
    \label{fig:supply-marble-compare}
\end{figure*}

\begin{figure*}
    \centering
    \includegraphics[width=0.85\linewidth]{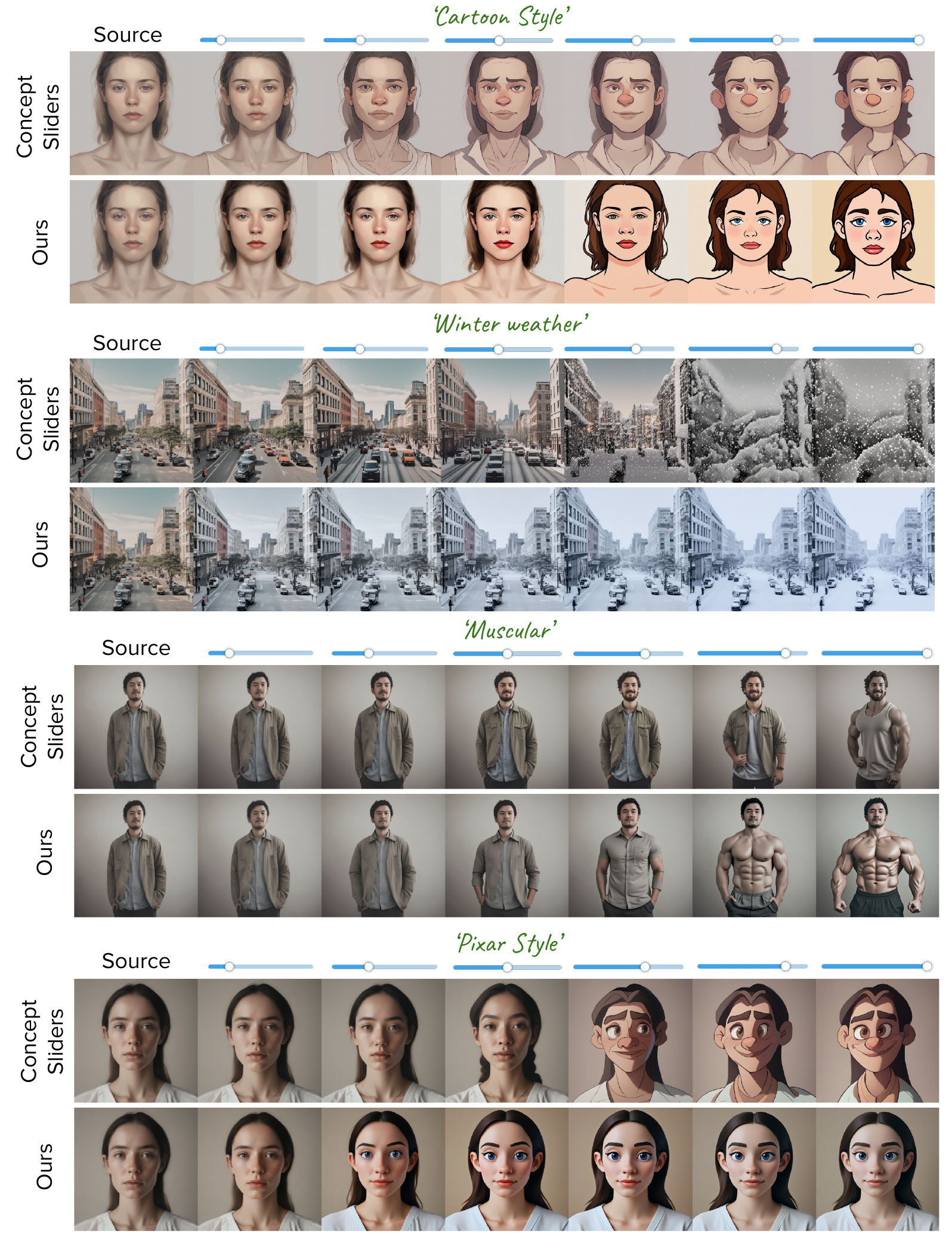}
    \caption{Comparison with Concept Sliders for diverse attribute editing.}
    \label{fig:supply-cslider-compare}
\end{figure*}

We present additional comparison with interpolation-based baselines in Fig.~\ref{fig:supply-baseline-1},~\ref{fig:supply-baseline-2} and with domain-specific methods ConceptSliders in Fig.~\ref{fig:supply-cslider-compare}, and MARBLE in Fig.~\ref{fig:supply-marble-compare}. 

\subsection{Additional Baselines}
\label{subsec:supply-baslines}
\begin{figure*}
    \centering
    \includegraphics[width=0.9\linewidth]{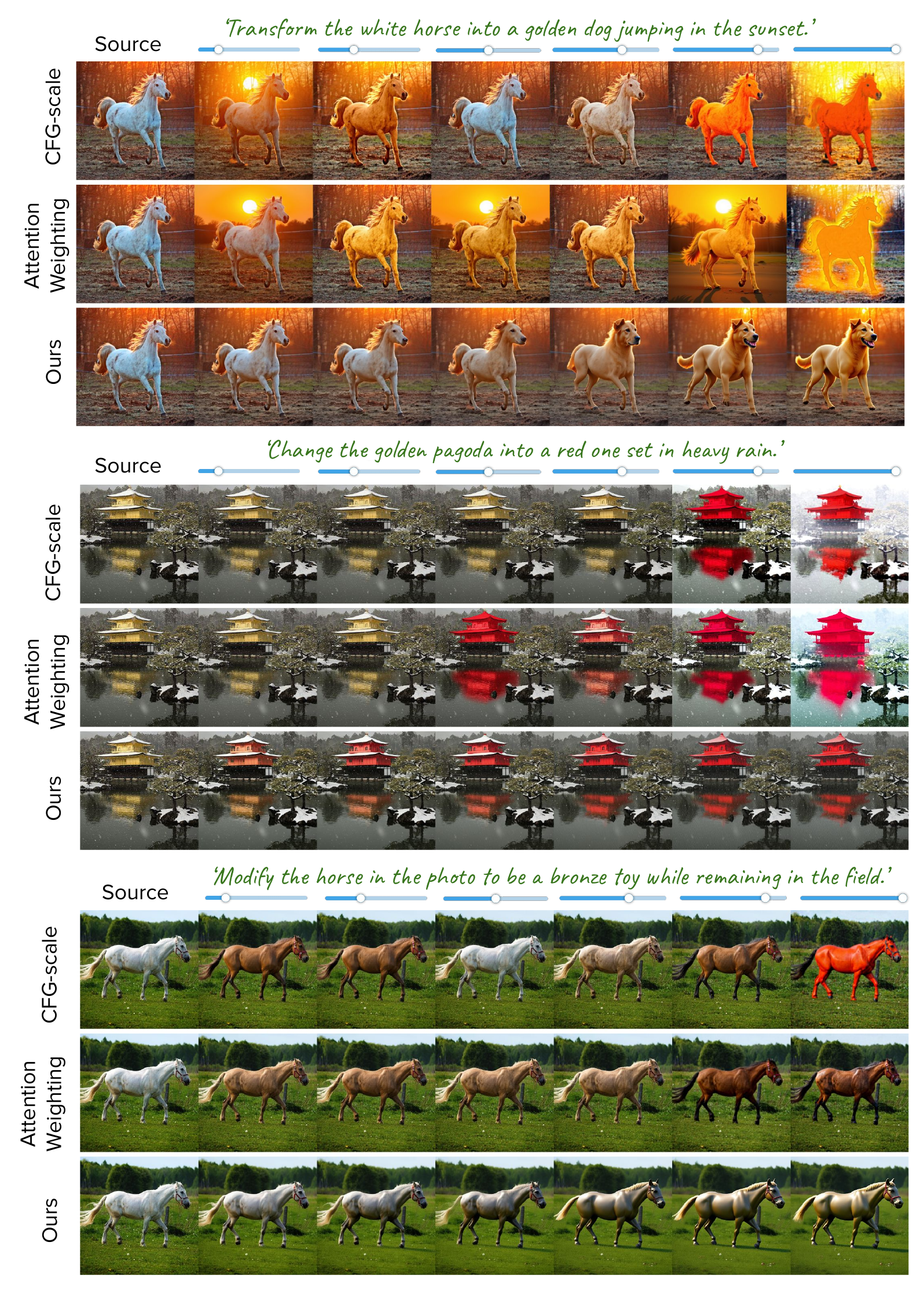}
    \caption{We compare with additional inference time baselines.}
    \label{fig:new-baselines}
\end{figure*}

We compared \emph{Kontinuous Kontext} with two additional simple baselines: a) CFG-Scale - We change the classifier free guidance scale to control the extent of the edit, as we expect that with a higher CFG scale, the generated edit should follow the edit instruction more closely. b) Attention reweighting - We scale the cross-attention maps between the text tokens and the generated visual tokens, inspired by Prompt2Prompt~\citep{hertz2022prompt}. The insight is that, if we increase the cross-attention weight between the target latents and the text instruction, the target image will pay more attention to the edit instruction, resulting in stronger edits. We present a comparison in Tab.~\ref{tab:inference-compare} and Fig.~\ref{fig:new-baselines}. Both the inference time baselines fail the generate smooth edit transitions and distort the input image identity significantly. These abrupt transitions are also evident as very high scores in $\delta_{smooth}$ smoothness metric.

\begin{table}[h]
\centering
\small
\resizebox{0.7\linewidth}{!}{
\begin{tabular}{@{}lcc@{}}
\toprule
Methods & $\delta_{\text{smooth}} \downarrow$ & CLIP-dir $\uparrow$ \\ \midrule
CFG-scale     & 152.205    &  \textbf{0.242}   \\ 
Attention-weighing     & 120.760   & 0.237    \\ 
Ours   & \textbf{0.329}    & 0.241   \\
\bottomrule 
\end{tabular}}
\vspace{-2mm}
\caption{\small Comparison with additional inference time baselines.}
\label{tab:inference-compare}
\end{table}

\subsection{Failure case - Extrapolation beyond the training strength $s>1$} 
\begin{figure}
    \centering
    \includegraphics[width=\linewidth]{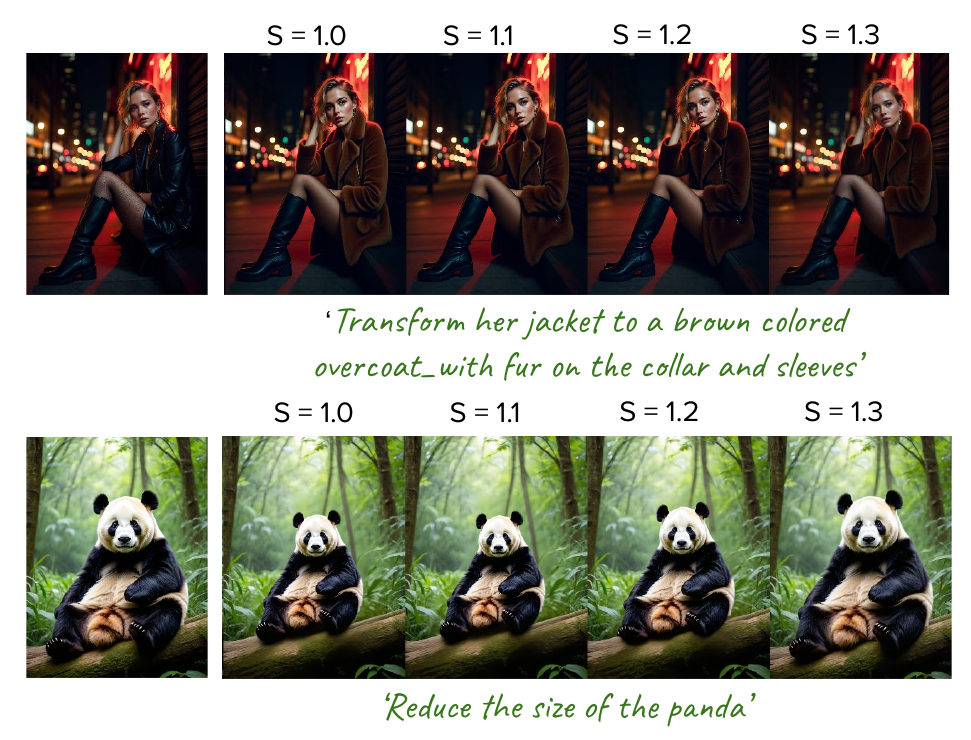}
    \caption{\textbf{Extrapolation of edit strengths}. One of the failure cases of our method is that it cannot generate edits with extrapolation well. In most cases, it either recreates the full edit image ($s=1$), or reduces the extent of edit in the extrapolation region.}
    \label{fig:extrapolation-samples}
\end{figure}

One of the failure cases of our method is in extrapolating edits beyond the strength value $s=1$. Our method either does not perform the edits for $s>1$ or reduces the extent of the edit as shown in Fig.~\ref{fig:extrapolation-samples}.

\subsection{ Evaluation dataset}
\textbf{PIEBench:} We evaluate our method on the widely used image editing benchmark PIEbench, which consists of diverse and challenging instruction-driven image editing test examples. The benchmark consists of edits from the following categories: change object, add/remove object, change pose, change color, change material, change background, and change style. We kept all edit categories except add/remove object, as it's a discrete edit, and our method focuses on continuous attribute editing. 

\vspace{1mm}
\noindent 
\textbf{Domain-specific comparison:} As ConceptSliders optimize for specific attributes (e.g., smile, chubby), we accordingly select the dataset suitable for such edits (e.g., face images) for comparison. Further, as concept-sliders is designed to achieve continuous control during image generation, and is not directly suitable for image editing, we evaluate it on $44$ generated images across $11$ sliders covering facial attributes, stylization, and scene edits. For material control, we evaluate MARBLE on $40$ images from the material editing category from the PIEBench dataset. 

\subsection{ Compositional Editing}
\begin{figure*}
\centering
\includegraphics[width=\linewidth]{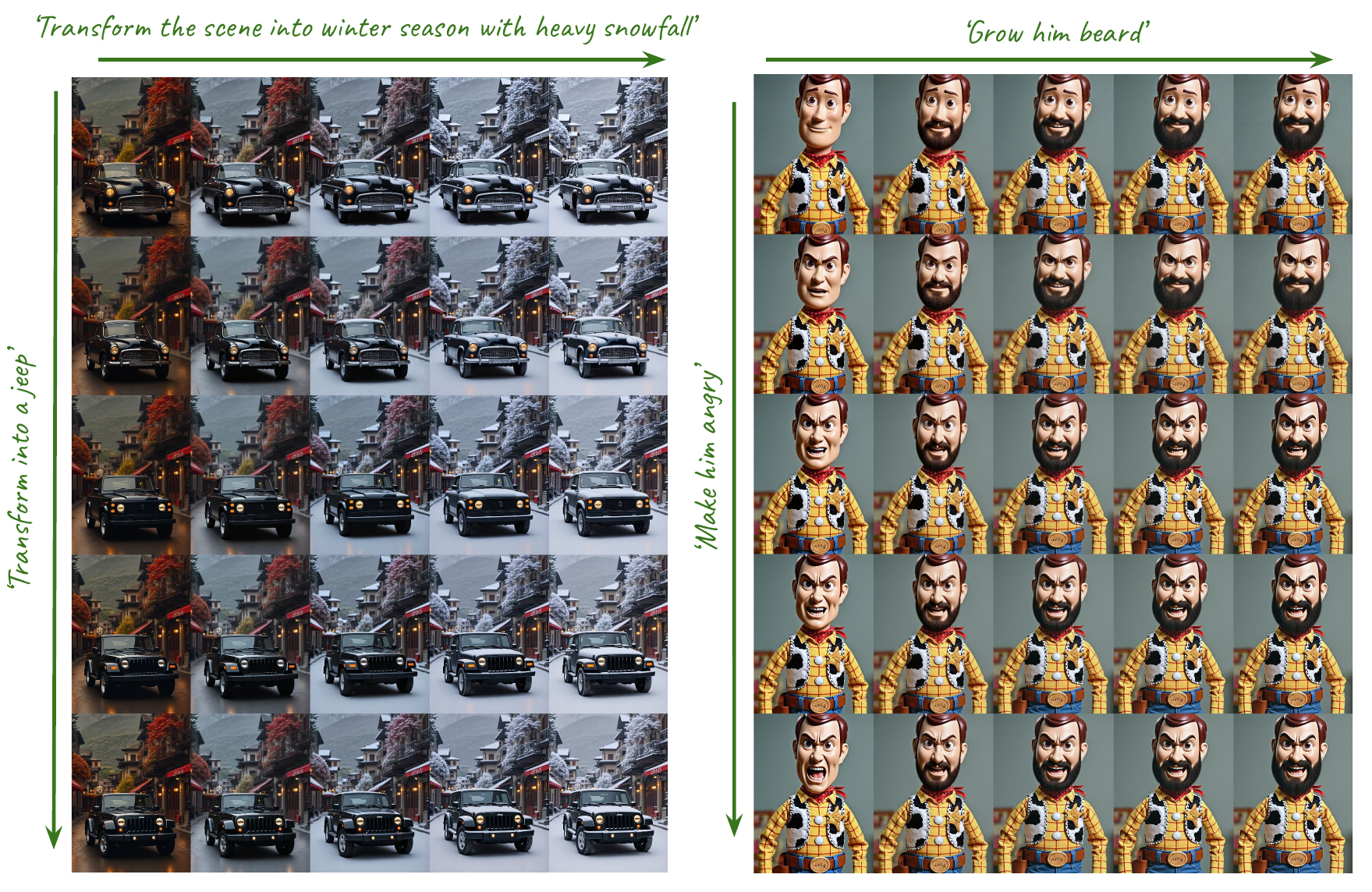}
\caption{Compositional editing with Kontinuous Kontext. We perform composition of two edits with fine-grained strength control for individual edits. We first generate the edits for the first prompt and then apply the second edit on the edited images from the first pass. Our method can generate realistic compositional edits, providing finer control for editing multiple attributes together.}
\label{fig:compositional-editing}
\end{figure*}

Kontinuous Kontext enables multi-turn editing like the base Flux Kontext model. We first edit the input image with the first edit instruction and followed by editing with the second edit instruction. The results are shown in Fig.~\ref{fig:compositional-editing}. 

\subsection{ Generalization to other architectures}
Our method builds on the findings~\citep{garibi2025tokenverse, dalva2024fluxspace} that the modulation space of recent DiT based diffusion models enables fine-grained control over the scene contents. Leveraging this observation, we train a strength-conditioned slider projector to predict the offsets for the modulation parameters of DiT blocks. Though we have demonstrated this capability with Flux Kontext, most of the recent instruction-driven image editing models consist of the same DiT blocks. Hence, our approach is compatible with all the recent instruction-based editing (e.g., ~\citep{wu2025qwen}) and can easily be integrated to achieve fine-grained strength control for image editing.

\begin{table}[]
\resizebox{\linewidth}{!}{
\begin{tabular}{@{}c|c|c@{}}
\toprule
Method         & $\#$ Diffusion Evaluations & Models used                    \\ \midrule
Diffmorpher    & N+1                   & Flux Kontext (1) + SD-2.1 (N)  \\
Freemorph      & N+1                   & Flux Kontext (1) + SD-2.1 (N)  \\
Wan            & 2                     & Flux Kontext (1) + WAN-2.1 (1) \\ \midrule
ConceptSliders & N                     & SD-XL (N)                      \\
MARBLE         & N                     & SD-XL (N)                      \\ \midrule
Ours           & N                     & Flux Kontext (N)               \\ \bottomrule
\end{tabular}}
\caption{\textbf{Inference cost comparison.} We compare against baselines in terms of the number of diffusion model evaluations, as the baselines use different base diffusion models. Here, $N$ represents the number of sequential edits generated for a given image and prompt. Our method uses the same or fewer inferences than all the baselines, except WAN, which requires a costly video model evaluation.}
\label{tab:baseline-latency}
\end{table}

\subsection{ Latency and Computational Cost}
\label{subsec:compute}
Our method adds negligible inference overhead over the base Flux Kontext model in terms of latency ($60s$ vs $56s$) on a single NVIDIA $A6000$ GPU for $28$ denoising steps. In comparison to the baselines, our method uses a comparable number of diffusion model inferences. As the baselines use different base diffusion models, we compare with them in terms of the number of diffusion evaluations in Table.~\ref{tab:baseline-latency}. Our trainable parameters include a small $4$ layer projector and LoRA parameters for the projection matrices of the base model, which accounts for $\approx1\%$ of parameters of the base Flux Kontext model.  

{
    \small
    \bibliographystyle{ieeenat_fullname}
    \bibliography{main}
}

\end{document}